\documentclass[10pt,twocolumn,letterpaper]{article}

\usepackage{iccv}

\usepackage{times}
\usepackage{epsfig}
\usepackage{graphicx}
\usepackage{amsmath}
\usepackage{amssymb}
\usepackage{mathtools}
\usepackage{siunitx}
\usepackage{booktabs}
\usepackage{pifont}

\usepackage{enumitem}
\usepackage{float}
\usepackage{multirow}
\usepackage{placeins}
\usepackage{titling}

\usepackage{caption}
\usepackage{subcaption}

\usepackage[dvipsnames]{xcolor}

\DeclareMathOperator*{\argmax}{arg\,max}

\newcommand{\cmark}{\color{ForestGreen}\ding{51}}
\newcommand{\xmark}{\color{BrickRed}\ding{55}}

\usepackage[pagebackref=true,breaklinks=true,colorlinks,bookmarks=false]{hyperref}

\def\iccvPaperID{10343}
\iccvfinalcopy

\begin{document}

\title{The Hitchhiker's Guide to Prior-Shift Adaptation}

\author{Tomas Sipka, Milan Sulc, Jiri Matas\\
Dept. of Cybernetics, Faculty of Electrical Engineering, Czech Technical University in Prague \\
 {\tt\small sipka3@seznam.cz, milansulc01@gmail.com, matas@fel.cvut.cz}
}

\date{}

\maketitle

\begin{abstract}
In many computer vision classification tasks, class priors at test time often differ from priors on the training set. In the case of such \textit{prior shift}, classifiers must be adapted correspondingly to maintain close to optimal performance. This paper analyzes methods for adaptation of probabilistic classifiers to new priors and for estimating new priors on an unlabeled test set. We propose a novel method to address a known issue of prior estimation methods based on confusion matrices, where inconsistent estimates of decision probabilities and confusion matrices lead to negative values in the estimated priors. Experiments on fine-grained image classification datasets provide insight into the best practice of prior shift estimation and classifier adaptation, and show that the proposed method achieves state-of-the-art results in prior adaptation. Applying the best practice to two tasks with naturally imbalanced priors, learning from web-crawled images and plant species classification, increased the recognition accuracy by 1.1\% and 3.4\% respectively.
\end{abstract}

{\let\thefootnote\relax\footnote{\hspace*{-5mm} Code is available at \url{https://github.com/sipkatom/The-Hitchhiker-s-Guide-to-Prior-Shift-Adaptation}.}}

\section{Introduction}
Let us consider probabilistic classifiers that estimate the posterior probability $p(Y|X)$, where $X$ and $Y$ are random variables describing an observation and its correct class label respectively. 
In the framework of empirical risk minimization, the classifier parameters are trained by minimizing the loss function on a training set, which is assumed to come from the same distribution as the expected test data. However, in many classification tasks, the class prior probabilities $p_\mathcal{E}(Y)$ on an evaluation set differ from $p_\mathcal{T}(Y)$ on the training set, while the class-conditional distributions remain unchanged, i.e. $p_\mathcal{E}(X|Y) =p_\mathcal{T}(X|Y)$. In case of
this phenomenon, called \textit{prior shift} or \textit{label shift}, classifiers require adaptation to maintain close to optimal performance.

For example, let us assume symptoms $X$ common to several diseases with fixed conditional probabilities $p(X|Y)$. If we use a previously trained classifier during an outbreak of a disease, the classification results should change according to the new prior probabilities $p_\mathcal{E}(Y)$. As another example, let us consider species classification, where specimens of a species have the same appearance model $p(X|Y)$, yet the species incidence $p_\mathcal{E}(Y)$ may change according to location, time of year, and other environmental factors.

Other domain adaptation scenarios considered in the literature include \textit{covariate shift} \cite{zadrozny2004learning,sugiyama2008direct}, also called \textit{sample selection bias} -- when the appearance model $p(X)$ changes, but the conditional output distribution $p(Y|X)$ is invariant; and \textit{conditional shift} \cite{zhang2013domain} -- where $p(Y)$ remains the same, but $p(X|Y)$ changes. This paper focuses solely on the problem of \textit{prior shift}, also denoted \textit{label shift} or \textit{target shift}.

Class priors often follow a long-tail (LT) distribution. Classification on the less frequent classes can be improved by specific imbalanced/long-tail data losses, such as Focal loss \cite{lin2017focal} or LDAM loss \cite{cao2019learning}, and training methods, such as OLTR \cite{liu2019large} or BLT \cite{kozerawski2020blt}. 
While we experiment with prior shifts from and to LT distributions, we focus on test-time adaptation of pre-trained classifiers with outputs approximating posterior probabilities, trained by cross entropy loss minimization. The paper does not aim at improving long-tail training methods, which is a different task than prior shift adaptation, where the shift can happen between arbitrary class distribution, not only long-tailed.  We consider the standard classification task, i.e. a Bayesian decision problem with a 0/1 loss. We thus aim at improving the accuracy of the pre-trained classifier after prior shift.

As the first contribution, the paper summarizes existing methods for adaptation to prior shift and experimentally answers questions about the best practice, such~as:
\begin{itemize}[noitemsep,topsep=0pt,leftmargin=*]
    \item Should an existing classifier be adapted to new priors by re-weighting its predictions, or is it worth it to re-train the classifier with training sampling matching the shift?
    \item What is the best practice to estimate, in an unsupervised way, class priors on a  test set?
    \item Is it better to directly estimate test priors, or shall the importance weights \cite{lipton2018detecting} be estimated?
    \item How does the estimate quality depend on the test set size?
\end{itemize}

As the second contribution, we propose a maximum likelihood approach for correcting all estimates based on the inversion of confusion matrix \cite{forman2008quantifying,saerens2002adjusting,vucetic2001classification}, where inconsistent estimates of decision probabilities and confusion matrices could result in negative values. Vucetic and Obradovic \cite{vucetic2001classification}, who use a bootstrapping framework, avoid such infeasible solutions by discarding corresponding bootstrap replicates.
McLachlan \cite{mclachlan1992discriminant} and Forman \cite{forman2008quantifying} mention clipping the estimate into the range 0–100\%. 
 To the best of our knowledge, this paper is the first to provide
 a well-grounded solution to this problem. The proposed method, (S)CM$^\text{L}$, achieves state-of-the-art results, in most cases performing better than existing methods including the ``Hard-To-Beat'' EM with Bias-Corrected Calibration \cite{alexandari2020maximum}.

As the third contribution, we propose a Maximum A-Posteriori estimation method (S)CM$^\text{M}$, extending the proposed CM-based likelihood maximization (S)CM$^\text{L}$ by adding a hyper-prior. We show that a Dirichlet hyper-prior, as in \cite{sulc2019improving}, improves the estimation of dense distributions, and performs better than the existing MAP estimation \cite{sulc2019improving}.

\section{Related Work}
\label{sec:related_work}
Several methods have been proposed to tackle adaptation to prior shift, either by adapting the predictions of a pre-trained classifier \cite{duplessis2012semi,saerens2002adjusting,sulc2019improving,vucetic2001classification} or by re-training the classifier with adjusted training sample weights \cite{azizzadenesheli2019regularized,lipton2018detecting}.

The new priors are commonly unknown, but can be estimated from an unlabeled set of observations. Vucetic and Obradovic \cite{vucetic2001classification} estimate the new priors $p_\mathcal{E}(Y)$ from the classifier's probabilities $p(D=i|Y=k)$ of predicting class $i$ when the true class is $k$, i.e. from the confusion matrix (CM) of the classifier.
Saerens et al. \cite{saerens2002adjusting} propose an EM algorithm for Maximum Likelihood Estimation (MLE) of priors from predictions of posterior probabilities $p(Y|X)$, and experimentally show an improvement compared to no adaptation and compared to a Confusion Matrix based estimate. Du Plessis and Sugiyama \cite{duplessis2012semi} prove that the EM procedure \cite{saerens2002adjusting} is equivalent to fixed-point-iteration minimization of the KL divergence between the new input density $p_\mathcal{E}(X)$ and the marginalization of the joint distribution $ \sum\limits_Y p_\mathcal{E}(Y) p(X|Y) $. Du Plessis and Sugiyama \cite{duplessis2012semi}
also propose methods for direct divergence minimization in cases where the input density $p(x)$ can be directly modeled, e.g. with a kernel-based non-parametric estimate. Sulc and Matas \cite{sulc2019improving} emphasize the importance of adapting to new priors in fine-grained image classification and propose a Maximum A Posteriori (MAP) estimation adding a Dirichlet hyper-prior.

Lipton et al. \cite{lipton2018detecting} propose a confusion matrix based estimate of the prior ratio $w(Y) = \frac{p_\mathcal{E}(Y)}{p_\mathcal{T}(Y)}$, forming a Black Box Shift Learning
(BBSL) framework. Azizzadenesheli et al. \cite{azizzadenesheli2019regularized} propose to increase the stability of the prior ratio estimation by regularizing the distribution shift, forming the Regularized Learning under Label
Shifts (RLLS) framework. Both methods \cite{azizzadenesheli2019regularized,lipton2018detecting} then re-train the classifier with sample weights according to the prior ratio.

Alexandri et al. \cite{alexandari2020maximum} propose a bias-corrected version of temperature scaling for classifier calibration and use the EM algorithm for prior estimation on the calibrated predictions. They experimentally show that adapting a calibrated classifier to new priors estimated by EM outperforms the re-trained classifiers using BBSL \cite{lipton2018detecting} and RLLS \cite{azizzadenesheli2019regularized}, making EM with Bias-Corrected Calibration ``Hard-To-Beat''.

In the following subsections, we assess the existing methods and formulate them in a unified notation.

\subsection{Classifier Adaptation}
Let $\mathbb X$ be a feature space, $K$ the number of classes and ${\mathbf f:\mathbb X \rightarrow \Delta_{K-1}}$ a classifier mapping observations $\mathbf x\in\mathbb X$ onto the probability simplex $\Delta_{K-1}$.
The classifier is trained to approximate class posteriors $f(\mathbf{x}) \approx p(Y|\mathbf{x})$, e.g. by cross-entropy minimization. The training set $\mathcal{T}=\{\mathbf{x}_i,y_i\}_{i=1}^{N}$ is sampled from distribution $p_\mathcal{T}(X,Y)$. In the case of \textit{prior shift}, the priors on evaluation set $\mathcal{E}=\{\mathbf{x}_i\}_{i=1}^{M}$ change to $p_\mathcal{E}(Y)$, while the appearance model $p_\mathcal{T} (X|Y) = p_\mathcal{E} (X|Y)$ remains the same.

We consider different cases of classifier adaptation, where the new priors $p_\mathcal{E}(Y)$ are either known or unknown, and the classifier:
\begin{enumerate}[noitemsep,topsep=0pt]
    \item is fixed and trained on a known training set $\mathcal{T}$,
    \item will be trained on the training set $\mathcal{T}$ and we can change the training procedure,
    \item is fixed and trained on an unknown training set $\mathcal{T}$.

\end{enumerate}

\subsection{Adaptation of a Fixed Classifier to New Priors}
A probabilistic classifiers $f_\mathcal{T} (\mathbf{x}) \approx p_\mathcal{T}(Y|\mathbf{x})$ is simply adapted  \cite{duplessis2012semi,saerens2002adjusting,sulc2019improving} to new a-priori probabilities $p_\mathcal{E}(Y)$   following the Bayes theorem:
\begin{equation}
     p_\mathcal{T}(\mathbf x|Y) = p_\mathcal{E}(\mathbf x|Y) = \frac{ p_\mathcal{T}(Y|\mathbf x) p_\mathcal{T}(\mathbf x)}{ p_\mathcal{T}(Y)} = \frac{ p_\mathcal{E}(Y|\mathbf x) p_\mathcal{E}(\mathbf x)}{ p_\mathcal{E}(Y)}
\end{equation}
The new predictive prior $ p_\mathcal{E}(Y|\mathbf x)$ is then:
\begin{equation}
     p_\mathcal{E}(Y|\mathbf x) =  p_\mathcal{T}(Y|\mathbf x)\frac{ p_\mathcal{E}(Y) p_\mathcal{T}(\mathbf x)}{ p_\mathcal{T}(Y) p_\mathcal{E}(\mathbf x)} \propto  p_\mathcal{T}(Y|\mathbf x)\frac{ p_\mathcal{E}(Y)}{ p_\mathcal{T}(Y)}
     \label{eq:adapt_priors}
\end{equation}

We can proceed with estimating $p_\mathcal{E}(Y)$ with one of the methods from Sections \ref{subsection:cm_prior_estimation} and  \ref{subsection:predictions_prior_estimation}. Alternatively, we can directly estimate the ratio of the priors $w(Y)= p_\mathcal{E}(Y)/ p_\mathcal{T}(Y)$ as in Section \ref{sec:prior_ratio}.

\subsection{Estimation of New Priors Based on Confusion Matrices}
\label{subsection:cm_prior_estimation}
A standard procedure \cite{mclachlan1992discriminant, saerens2002adjusting} for prior estimation is based on a $K\times K$ confusion matrix (CM) in the format $\mathbf{C}_{d|y}$, where the value in the $k$-th column and $i$-th row is the probability $p(D=i|Y=k)$ of classifier $\mathbf f$ deciding for class $i$ when the true class is $k$. Assuming that the density $p_\mathcal{T} (X|Y) = p_\mathcal{E} (X|Y)$  remains unchanged, the confusion matrix $\mathbf C_{d|y}$ of a classifier does not change with prior shift \cite{lipton2018detecting}. 
 Marginalizing over the joint density $p(D,Y)$:\vspace*{-1.0em}\\
\begin{equation}\label{eq:cm_conditional_marginalization}
    \begin{split}
        p(D=i)&=\sum_{k=1}^K p(D=i|Y=k)p(Y=k) \\
        p(D) &= \mathbf C_{d|y} p(Y)
    \end{split}
\end{equation}
McLachnan \cite{mclachlan1992discriminant} and Saerens et al. \cite{saerens2002adjusting} simply compute the new priors $p_\mathcal{E}(Y)$ from Equation \eqref{eq:cm_conditional_marginalization}:
\begin{equation}\label{eq:cm_computing_priors}
    \hat p_\mathcal{E}(Y) = \mathbf{\hat C}_{d|y}^{-1}\hat p_\mathcal{E}(D),
\end{equation}
using an estimate of $\mathbf C_{d|y}$ computed on a validation set and an estimate of $p(D)$ computed by counting the classifier decisions on the test set.

Let us also consider a \textit{soft confusion matrix}\footnote{Following the terminology of Lipton et al. \cite{lipton2018detecting}.} (SCM)  $\mathbf C_{d|y}^\text{soft}$  estimated from the classifier's soft predictions $\mathbf{f}$ as
\begin{equation}\label{eq:soft_estimate}
    \mathbf{\hat c}_{:,k}^\text{soft} =\frac{1}{N_k} \sum_{\mathbf{x}_i:y_i=k} \mathbf{f}(\mathbf{x}_i),
\end{equation}
where $\mathbf{\hat c}_{:,k}^{\text{soft}}$ denotes the \textit{k}-th column of SCM. The probability $p_\mathcal{E}^\text{soft}(D)$ can be estimated by averaging predictions $f(\mathbf{x})$ over the test set. The new priors are then computed similarly to Equation \eqref{eq:cm_computing_priors}.

\subsection{Estimation of New Priors Based on Posterior Predictions}
\label{subsection:predictions_prior_estimation}

\subsubsection{Maximum Likelihood and EM Algorithm}
\label{sec:mle_em}
Saerens et al. \cite{saerens2002adjusting} suggested to estimate priors  $p_\mathcal{E}(Y)$ on the evaluation set $\mathcal{E}=\{\mathbf{x}_i\}^M_{i=1}$ by maximizing the likelihood:
\begin{equation}
\vspace*{-1em}
    L(\mathcal{E})=\prod_{i=1}^{M}p_\mathcal{E}(\mathbf x_i)
    =\prod_{i=1}^{M}\sum_{k=1}^{K}p_\mathcal{E}(\mathbf x_i|Y=k)p_\mathcal{E}(Y=k)
\end{equation}
\vspace*{-0.5em}

They proposed an EM algorithm which iteratively re-computes the prior estimates:
\vspace*{-0.5em}\\
\begin{equation} \label{eq:exp_step}
    \hat p^{(s)}_\mathcal{E}(Y=k|\mathbf x_i) \coloneqq \frac{\hat p_\mathcal{T}(Y=k|\mathbf x_i)\frac{\hat p^{(s)}_\mathcal{E}(Y=k)}{\hat p_\mathcal{T}(Y=k)}}{\sum_{j=1}^{N'} \hat p_\mathcal{T}(Y=j|\mathbf x_i)\frac{\hat p^{(s)}_\mathcal{E}(Y=j)}{\hat p_\mathcal{T}(Y=j)}}
\end{equation}
\vspace*{-0.5em}
\begin{equation} \label{eq:max_step}
    \hat p_\mathcal{E}^{(s+1)}(Y=k) \coloneqq \frac{1}{M}\sum_{i=1}^{M} \hat p^{(s)}_\mathcal{E}(Y=k|\mathbf x_i),
\end{equation}
where $\hat p_\mathcal{T}(Y|\mathbf x_i)$ is modeled by the classifier's output $\mathbf{f}(\mathbf{x}_i)$.

Du Plessis and Sugiyama \cite{duplessis2012semi} showed that the EM algorithm can be derived from minimization of KL divergence between $p_\mathcal{E}(\mathbf{x})$ and its approximation. This leads to maximization of log-likelihood $l(\mathcal{E})=\log L(\mathcal{E})$:
\begin{equation}
    \begin{split}
        \mathbf{\hat P^*}=\argmax_{\mathbf{\hat P}} \frac{1}{M}\sum_{i=1}^{M}\log \sum_{k=1}^{K} \hat P_k \frac{p_\mathcal{T}(Y=k|\mathbf x_i)}{p_\mathcal{T}(Y=k)} \\
        s.t. \sum_{k=1}^K \hat P_k = 1; \quad \forall k :\hat P_k \ge 0,
    \end{split}
\end{equation}\vspace*{-1.0em}
\\where $\hat P_k=\hat p_\mathcal{E}(Y=k)$.

Sulc and Matas \cite{sulc2019improving} experimented with maximizing the log-likelihood function by projected gradient ascent:\vspace*{-0.5em}\\
\begin{equation}\label{eq:pgd_likelihood}
    \hat P_k^{s+1}=\pi\left(\hat P_k^{s}+\frac{\partial l(\mathcal E)}{\partial \hat P_k}\right)
\end{equation} \vspace*{-0.5em}\\
where $\pi(\cdot)$ denotes projection onto the probability simplex and
\begin{equation}
    \frac{\partial l(\mathcal E)}{\partial \hat P_k}=\sum_{i=1}^{M} \frac{\frac{p_\mathcal{T}(Y=k|\mathbf x_i)}{p_\mathcal{T}(Y=k)}}{\sum_{j=1}^{K} \hat P_k \frac{p_\mathcal{T}(Y=j|\mathbf x'_i)}{p_\mathcal{T}(Y=j)}}.
\end{equation}
The experimental results showed that the EM algorithm converged faster than gradient ascent, while achieving similar results.
\vspace*{-2.0em}\\
\subsubsection{Maximum Aposteriori Estimation}
Sulc and Matas \cite{sulc2019improving} proposed a maximum a-posteriori estimation:
\begin{equation}
    \begin{split}
        \mathbf{\hat P^*} = \argmax_{\mathbf{\hat P}} p(\mathbf{\hat P}|\mathcal E) = \argmax_{\mathbf{\hat P}} p(\mathbf{\hat P})p(\mathcal E|\mathbf{\hat P}) \\
        =\argmax_{\mathbf{\hat P}} \log p(\mathbf{\hat P})+ \underbrace{ \log p(\mathcal E|\mathbf{\hat P}) }_{l(\mathcal{E})}
    \end{split}
\end{equation}
where the distribution $p(\mathbf{\hat P})$ is a hyper-prior representing some additional knowledge about class distribution. Specifically, they used a symmetric Dirichlet distribution $\mathrm{Dir}(\boldsymbol{\alpha})$.

The solution to maximum a-posteriori can be found by projected gradient ascent, adding the derivative of $\log \mathrm{Dir}(\boldsymbol{\alpha})$ into the $\pi(\cdot)$ function: in Equation \eqref{eq:pgd_likelihood}:

\begin{equation} \label{eq:dirichlet_derivative}
    \frac{\partial \log p(\mathbf{P})}{\partial P_k} =  \frac{\partial \log \mathrm{Dir}(\boldsymbol{\alpha})} {\partial P_k} = \frac{\alpha -1}{ P_k}
\end{equation}

\subsection{Estimation of Prior Ratio Based on Confusion Matrices}
\label{sec:prior_ratio}
Lipton et al. \cite{lipton2018detecting} estimate the prior ratio $w(Y)= \frac{p_\mathcal{E}(Y)}{p_\mathcal{T}(Y)}$ using a confusion matrix in the format $\mathbf{C}_{d,y}$, i.e. with joint probability $p(D=i,Y=k)$, unlike the conditional probability used in Section \ref{subsection:cm_prior_estimation}. Since ${p_\mathcal{T} (D|Y) = p_\mathcal{E}(D|Y)}$:
\begin{equation}\label{cm_ratio_estimation}
    \begin{split}
        p_\mathcal{E}(D=i)= & \sum_{k=1}^K p_\mathcal{T}(D=i|Y=k)p_\mathcal{E}(Y=k) = \\
        = & \sum_{k=1}^K p_\mathcal{T}(D=i,Y=k) \underbrace{\frac{p_\mathcal{E}(Y=k)}{p_\mathcal{T}(Y=k)}}_{w(Y=k)}\\
        p_\mathcal{E}(D) = &\; \mathbf{C}_{d,y} w(Y) \implies \hat w(Y) = \mathbf{\hat C}_{d,y}^{-1}\hat p_\mathcal{E}(D)
    \end{split}
\end{equation}

This estimation is called Black Box Shift Estimation (BBSE). A variant using a soft confusion matrix $\mathbf C_{d,y}^\text{soft}$  is denoted BBSE-S.

Note that even the estimation of prior ratio may suffer from inconsistent estimates of the confusion matrix and $p(D)$, as demonstrated in the Supplementary Material.

\subsection{Classifier Calibration}
In the aforementioned methods, we often treated the classifier outputs $f_\mathcal{T}(\mathbf{x})$ as estimates of posterior probability $p_\mathcal{T}(Y|\mathbf{x})$. In practice, outputs of common probabilistic classifiers, such as Convolutional Neural Networks, tend to provide over-confident predictions due to over-fitting to the training set. Guo et al. \cite{guo2017calibration} study confidence calibration in the context of neural networks, and compare several models for classifier calibration, of which a simple temperature scaling (TS) procedure performs the best in terms of the calibration error. With temperature scaling, the softmax logits $z(\mathbf{x})$ are divided by the temperature $T$:

\begin{equation}
    p^\text{TS}(y=i | \mathbf{x}) = \dfrac{\exp\left(z_i(\mathbf{x}) / T \right)}{\sum_j \exp\left(z_i(\mathbf{x}) / T \right)}
\end{equation}

While lowering the calibration error, Alexandri et al. \cite{alexandari2020maximum} show that temperature scaling is not a suitable calibration for adaptation to prior shift, possibly because of large systematic biases in the calibrated probabilities. They propose Bias-Corrected Temperature Scaling (BCTS), adding a class-specific bias term:

\begin{equation}
    p^\text{BCTS}(y=i | \mathbf{x}) = \dfrac{\exp(z_i(\mathbf{x}) / T + b_i)}{\sum_j \exp(z_j(\mathbf{x}) / T + b_j)}
\end{equation}

Alexandri et al. \cite{alexandari2020maximum} show that such bias-corrected classifier calibration improves prior-adaptation with the EM-algorithm \cite{saerens2002adjusting} from Section \ref{sec:mle_em}, outperforming both BBSL \cite{lipton2018detecting} and RLLS \cite{azizzadenesheli2019regularized}.

\subsection{Training Data Sampling Strategies for Prior Shift Adaptation}
\label{sec:sampling_strategies}
A possible alternative to adapting the predictions $f(\mathbf{x}) \approx p_\mathcal{T}(Y|\mathbf{x})$ following Equation \eqref{eq:adapt_priors} is to instead train a new classifier $f_\mathcal{E}(\mathbf{x}) \approx p_\mathcal{E}(Y|\mathbf{x})$ by changing the sampling strategy from the training set according to $p_\mathcal{E}(Y)$. A similar approach was used e.g. in the winning submission of iNaturalist 2017 \cite{cui2018large}, where the training data was highly imbalanced, while the validation and test data were rather balanced in terms of class priors. The classifier in \cite{cui2018large} was first trained on the full training set and then fine-tuned on a balanced subset of the training set. Unlike \cite{cui2018large}, we propose to use all training examples, but sample the training data following  $p_\mathcal{E}(Y)$.

Prior adaptation following Equation \eqref{eq:adapt_priors} will be compared to re-training the network with adapted sampling in the experiments in Section \ref{sec:exp_adapt_retrain}. The practical disadvantage of the later approach is clear: the necessity to re-train the classifier with every new prior distribution. 

\section{Proposed Methods}

\subsection{Maximum Likelihood Estimate Based on Confusion Matrices}
\label{subsubsection:cm_maximum_likelihood}

As observed in the literature \cite{forman2008quantifying,mclachlan1992discriminant, vucetic2001classification}, in some cases Equation \eqref{eq:cm_computing_priors} can result in a vector outside of the $\Delta_{K-1}$ simplex, i.e. the estimate can contain negative values. We observe this phenomenon is common in practice. 

\begin{figure}[tb]
\centering
\includegraphics[width=.45\textwidth]{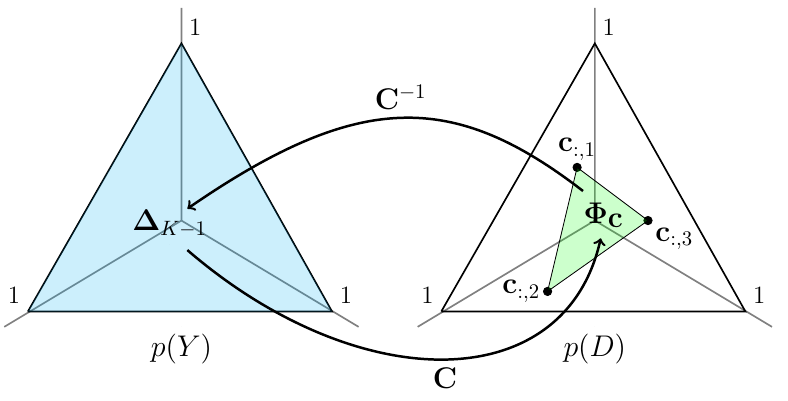}
\vspace*{-0.5em}
\caption{The convex set $ \Phi_{\mathbf{C}} \subset \Delta_{K-1} $ of all possible values of $p(D)$ for a classifier with confusion matrix $\mathbf{C}$.}
\label{fig:simplex_subset}
\vspace*{-0.5em}
\end{figure}

Following Equation \eqref{eq:cm_conditional_marginalization}, the probability of classifier decisions $p(D)$ is a convex combination of columns in $\mathbf{C}_{d|y}$ as $p(Y) \in \Delta_{K-1}$. Since the columns of the confusion matrix are probability vectors, they define a convex set $\Phi_{\mathbf{C}}$ of feasible values $p(D)$ within the probability simplex $\Delta_{K-1}$. 
In other words, a classifier with confusion matrix $\mathbf{C}$  will result in decisions from $p(D) \in \Phi_{\mathbf{C}}$.  The class distribution $p(Y)$ determines the value of $p(D)$ within $\Phi_{\mathbf{C}}$. See Figure \ref{fig:simplex_subset} for illustration.
For the true distribution $p(D)$ and confusion matrix $\mathbf{C}_{d|y}$, Equation \eqref{eq:cm_conditional_marginalization} holds. The problem occurs when we work with estimates of the true distribution $\hat p(D)$ and confusion matrix $\mathbf{\hat{C}}_{d|y}$. If the estimates computed from a limited sample are not consistent, there may be no prior probability $\hat{p}(Y)$ satisfying Equation \eqref{eq:cm_conditional_marginalization}: For example, having 
$\mathbf{\hat{C}}_{d|y}=\begin{bmatrix}0.8 & 0.2\\ 0.2 & 0.8\end{bmatrix}, \; \hat p(D) = \begin{bmatrix}1 \\ 0\end{bmatrix},$
the unique solution to Equation \eqref{eq:cm_conditional_marginalization} is $\hat p(Y) = \begin{bmatrix} \frac{4}{3} \\ -\frac{1}{3}\end{bmatrix}$.

We propose a novel procedure for prior estimation based on maximizing the likelihood of $p_\mathcal{E}(D)$, which handles inconsistent estimates of $\hat p(D)$ and confusion matrix $\mathbf{\hat{C}}_{d|y}$, as can work even with singular matrices $\mathbf{\hat{C}}_{d|y}$, as it does not use matrix inversion.

Let $\mathbf{n}=(n_1, \ldots, n_K)$ be the numbers of classifier's decisions for class $1,\ldots, K$ on test set $\mathcal{E}$ and let us denote 
$ \mathbf{Q} = (q_1, \ldots, q_k) \coloneqq p_\mathcal{E}(D)$ the probabilities of classifier decisions on the test  distribution $p_\mathcal{E}(X,Y)$.
Assuming the independence of classifier decisions on the test set $\mathcal{E}$, the likelihood of $\mathbf{Q}$ follows by the multinomial distribution:

\begin{equation}
    L(\mathbf{Q}) = p(\mathbf{n}|\mathbf{Q}) = \frac{(n_1+\ldots+n_K)!}{n_1! \cdot \ldots \cdot n_K!} \cdot q_1^{n_1} \cdot \ldots \cdot q_K^{n_K}
\end{equation}

Substituting Equation \eqref{eq:cm_conditional_marginalization} into the likelihood function $L(\mathbf{Q})$, we can express the likelihood function of class priors $\mathbf{P}$:
\begin{equation}
    L(\mathbf{P}) = p(\mathbf{n}|\mathbf{P}) = \frac{(n_1+\ldots+n_K)!}{n_1! \cdot \ldots \cdot n_K!}
    \prod\limits_{k=1}^K ( \mathbf{c}_{k,:} \cdot \mathbf{P} )^{n_k},
\end{equation}
where $\mathbf{c}_{k,:}$ is the  $k$-th row of $\mathbf{C}_{d|y}$.

The log-likelihood is:
\begin{equation}
\begin{split}
 \ell(\mathbf{P}) = \log p(n|\mathbf{P}) =  \sum_{k=1}^K n_k\log ( \mathbf{c}_{k,:} \cdot \mathbf{P} ) + \theta_\mathbf{n},
\end{split}
\label{eq:cm_likelihood}
\end{equation}
where $\theta_\mathbf{n}$ is constant for a fixed $\mathbf{n}$.

We estimate the new class priors by maximizing the log-likelihood from Equation  \eqref{eq:cm_likelihood}:
\begin{subequations}\label{eq:cm_max_problem}
    \begin{align} 
        \mathbf{\hat P} = &\argmax_{\mathbf{P}} \ell(\mathbf{P}) =\argmax_{\mathbf{P}}\sum_{k=1}^K n_k\log \mathbf{c}_{k,:}\mathbf{P}\label{eq:cm_likelihood_criterion}\\
        \text{s.t.:} & \quad \sum_{k=1}^K P_k = 1; \quad \forall k : P_k \ge 0\label{eq:cm_likelihood_constraint}
    \end{align}
\end{subequations}
The convex objective can be iteratively maximized using projected gradient ascent:
\begin{equation}\label{eq:cm_pgd}
    \mathbf{\hat P}^{s+1}=\pi\left(\mathbf{\hat P}^{s} +  \nabla \ell(\mathbf{P}^s) \right) 
\end{equation}
where $\pi(\cdot)$ denotes projection onto the probability simplex  \cite{wang2013projection} and the gradient is computed as:
\begin{equation} \label{eq:cm_likelihood_derivative}
    \nabla \ell(\mathbf{P}) =\sum_{k=1}^{K}\frac{n_k}{\mathbf{c}_{k,:}\cdot\mathbf{P}}\mathbf{c}_{k,:}
\end{equation}

\subsection{Maximum A Posteriori Estimate Based on Confusion Matrices}
\label{subsubsection:cm_maximum_aposteriori}
Additional assumptions on the distribution $\mathbf{P}$ can be formulated as a hyper-prior $p(\mathbf{P})$. We can then extend the proposed procedure from Section \ref{subsubsection:cm_maximum_likelihood} to formulate maximum a-posteriori (MAP) estimation:
\begin{equation}\label{maximum_aposteriori_CM}
    \begin{split}
        \mathbf{\hat P_{MAP}} &= \argmax_{\mathbf{P}} p(\mathbf{P}|\mathbf n) = \argmax_{\mathbf{P}} p(\mathbf{P})p(\mathbf n|\mathbf{P}) \\
        &=\argmax_{\mathbf{P}} \log p(\mathbf{P})+ \argmax_{\mathbf{P}} \log p(\mathbf n|\mathbf{P}) \\
        \text{s.t.:} \quad & \forall k : P_k \ge 0; \quad \sum_{k=1}^K P_k = 1
    \end{split}
\end{equation}

where $p(\mathbf{P})$ denotes a hyper-prior on $\mathbf{P}$ and $\log p(\mathbf n|\mathbf{P})$ is log-likelihood given by Equation \eqref{eq:cm_likelihood}. 

Following \cite{sulc2019improving} we use a symmetric Dirichlet hyper-prior $\mathrm{Dir}(\alpha)$,
favouring dense distributions $\mathbf{P}$ with $\alpha>1$, and a sparse distribution for $0<\alpha<1$.

The solution to maximum a-posteriori can be found by projected gradient ascent, adding the hyper-prior derivative from Equation \eqref{eq:dirichlet_derivative} into the $\pi(\cdot)$ function in Equation \eqref{eq:cm_pgd}.

\section{Experiments}

In this section, we compare the existing and proposed methods for prior shift adaptation on existing long-tailed versions of standard image classification datasets: the CIFAR100-LT \cite{cao2019learning}, Places365-LT \cite{liu2019large} and ImageNet-LT \cite{liu2019large}.
Unlike Cao et al. \cite{cao2019learning}, our experiments require a validation set. Therefore, our training set, denoted as CIFAR100-LT$\*$, is smaller than of the original CIFAR100-LT, keeping 50 samples from each class for the validation set. Using the same script as Cao et al. \cite{cao2019learning} to sample the training set, the resulting imbalance ratio of 112.5 slightly differs from the original ratio of 100. For Places365-LT and ImageNet-LT, we use the same training and validation splits as Liu et al. \cite{liu2019large}. Networks trained on these long-tailed datasets are then evaluated on uniformly distributed test sets (UNI).
We also provide experiments in the other direction, denoted as UNI$\rightarrow$LT, where  networks trained on the full CIFAR100 and Places365 datasets are evaluated on test sets subsampled from the full test sets following the prior distributions of CIFAR100-LT and Places365-LT. 
Additional experiments on subsets of CIFAR100 and Places365 with hand-picked class distributions are in the suppl. material.

\begin{figure*}[t]
    \centering
 	\includegraphics[width=1.00\textwidth]{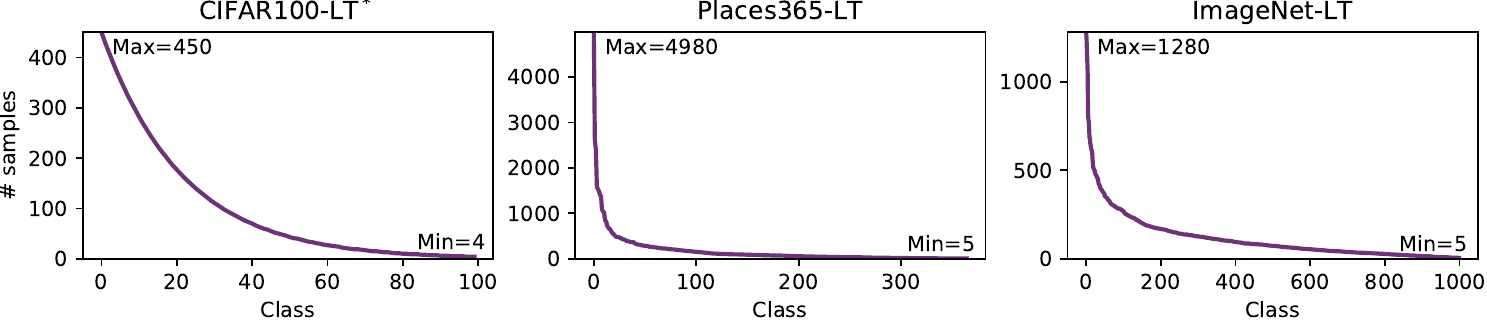}
 	\vspace*{-1.5em}
 	\caption{Long-tailed class distributions used in the CIFAR100-LT$^*$ \cite{cao2019learning}, Places365-LT \cite{liu2019large} and ImageNet-LT \cite{liu2019large} datasets. Note that our CIFAR-100-LT$^*$ slightly differs from the original CIFAR-100-LT \cite{cao2019learning}, which did not have a validation set.}
 	\label{fig:lt_distribution}
\end{figure*}

 \begin{table*}[hbt] 
 \centering
 \setlength\tabcolsep{3pt}
 \hspace*{-1.55mm}
 \begin{tabular}{m{1.6cm}|m{1.35cm}||>{\centering\arraybackslash}m{1.35cm}||>{\centering\arraybackslash}m{1.9cm}>{\centering\arraybackslash}m{1.9cm}>{\centering\arraybackslash}m{1.9cm}|>{\centering\arraybackslash}m{6.15cm}}
 \multicolumn{3}{c||}{}  & \multicolumn{3}{c|}{Standard training sampler} & Sampler follows $p_\mathcal{E}$ \\
 \multicolumn{2}{c||}{Dataset} &  \shortstack{BCTS \cite{alexandari2020maximum}\\ calibrated} & NA & $ \dfrac{p_\mathcal{E}}{\hat p^N_\mathcal{T}} $  &
 $ \dfrac{p_\mathcal{E}}{\hat p^{f}_\mathcal{T}} $  &  NA \\ 

 \hline \hline
  \multirow{4}{*}{\parbox{1.6cm}{\centering CIFAR100$^*$}} &
 LT$\rightarrow$UNI & \xmark  &  31.66$^{\scalebox{0.6}{±1.27}}$ & 33.99$^{\scalebox{0.6}{±1.41}}$ &  \textbf{34.06$^{\scalebox{0.6}{±1.35}}$} & 22.44 \\
 & LT$\rightarrow$UNI & \cmark  &  31.71$^{\scalebox{0.6}{±1.29}}$ & 30.41$^{\scalebox{0.6}{±1.57}}$ &  \textbf{34.54$^{\scalebox{0.6}{±1.32}}$} & 22.44 \\
 & UNI$\rightarrow$LT & \xmark &  63.83$^{\scalebox{0.6}{±0.82}}$ & \textbf{69.14$^{\scalebox{0.6}{±0.61}}$} & 69.13$^{\scalebox{0.6}{±0.58}}$ & 67.34 \\
 & UNI$\rightarrow$LT & \cmark &  63.83$^{\scalebox{0.6}{±0.82}}$ & 70.63$^{\scalebox{0.6}{±0.75}}$ &  \textbf{70.65$^{\scalebox{0.6}{±0.75}}$} & 67.34 \\

 \hline
 \multirow{2}{*}{\parbox{1.5cm}{\centering Places365}} 
 & LT$\rightarrow$UNI & \xmark  &  25.14$^{\scalebox{0.6}{±0.14}}$ & 28.03$^{\scalebox{0.6}{±6.09}}$ & \textbf{32.99$^{\scalebox{0.6}{±0.46}}$} & 24.98 \\
 & LT$\rightarrow$UNI & \cmark  &  25.16$^{\scalebox{0.6}{±0.14}}$ & 27.36$^{\scalebox{0.6}{±6.18}}$ &  \textbf{33.51$^{\scalebox{0.6}{±0.25}}$} & 25.00 \\

 \hline
   \multirow{2}{*}{\parbox{1.5cm}{\centering ImageNet}}
  & LT$\rightarrow$UNI & \xmark &  34.30$^{\scalebox{0.6}{±0.19}}$ & \textbf{37.36$^{\scalebox{0.6}{±0.07}}$} &  37.34$^{\scalebox{0.6}{±0.15}}$ & 30.01 \\
  & LT$\rightarrow$UNI & \cmark &  34.31$^{\scalebox{0.6}{±0.19}}$ & 36.07$^{\scalebox{0.6}{±0.30}}$ &  \textbf{37.45$^{\scalebox{0.6}{±0.19}}$} & 30.01 \\
 \end{tabular}
 \caption{\textit{``Adapt or Re-Train?''} Accuracy (± std. dev.) of classifiers adapted to new known priors $p_\mathcal{E}(Y)$ with different estimates of trained priors ($\hat p_\mathcal{T}^N$,$\hat p_\mathcal{T}^f$), compared to training a classifier with a sampling strategy following $p_\mathcal{E}(Y)$. NA denotes no adaptation of predictions. Results of classifier adaptation on CIFAR are averaged from 10 experiments, on Places365 and ImagNet from 5 experiments respectively. Re-training the classifier with a sampler following $p_\mathcal{E}(Y)$ was only experimented once for each dataset.}
 \label{tab:adapt_or_retrain_cifar_pt} 
 \end{table*}

\begin{table*}[hbt] 
\centering
\setlength\tabcolsep{3pt}
\hspace*{-1.55mm}\begin{tabular}{m{1.6cm}|m{1.35cm}||>{\centering\arraybackslash}m{1.35cm}||>{\centering\arraybackslash}m{1.90cm}|>{\centering\arraybackslash}m{1.90cm}>{\centering\arraybackslash}m{1.90cm}|>{\centering\arraybackslash}m{1.90cm}>{\centering\arraybackslash}m{1.90cm}|>{\centering\arraybackslash}m{1.90cm}}
\multicolumn{2}{c||}{Dataset} & \shortstack{BCTS \cite{alexandari2020maximum}\\ calibrated} & NA & CM  & CM$^\text{L}$  & SCM  & SCM$^\text{L}$ & Oracle \\
\hline \hline
 \multirow{4}{*}{\parbox{1.6cm}{\centering CIFAR100$^*$}}
& LT$\rightarrow$UNI & \xmark &  31.66$^{\scalebox{0.6}{±1.27}}$ &  21.37$^{\scalebox{0.6}{±2.68}}$ &  \textbf{33.00$^{\scalebox{0.6}{±1.56}}$} &           26.64$^{\scalebox{0.6}{±3.85}}$ &  \textbf{33.47$^{\scalebox{0.6}{±1.33}}$} &  34.06$^{\scalebox{0.6}{±1.35}}$ \\
& LT$\rightarrow$UNI & \cmark &  31.67$^{\scalebox{0.6}{±1.27}}$ &  19.11$^{\scalebox{0.6}{±2.98}}$ &  \textbf{32.41$^{\scalebox{0.6}{±1.50}}$} &           26.98$^{\scalebox{0.6}{±3.76}}$ &  \textbf{33.42$^{\scalebox{0.6}{±1.51}}$} &  34.40$^{\scalebox{0.6}{±1.38}}$ \\
& UNI$\rightarrow$LT & \xmark &  63.83$^{\scalebox{0.6}{±0.82}}$ &  68.06$^{\scalebox{0.6}{±0.92}}$ &  \textbf{68.08$^{\scalebox{0.6}{±0.75}}$} &           68.10$^{\scalebox{0.6}{±0.81}}$ &  \textbf{68.24$^{\scalebox{0.6}{±0.75}}$} &  69.13$^{\scalebox{0.6}{±0.58}}$ \\
& UNI$\rightarrow$LT & \cmark &  63.83$^{\scalebox{0.6}{±0.82}}$ &  69.08$^{\scalebox{0.6}{±0.94}}$ &  \textbf{69.10$^{\scalebox{0.6}{±0.98}}$} &           69.31$^{\scalebox{0.6}{±0.94}}$ &  \textbf{69.40$^{\scalebox{0.6}{±0.77}}$} &  70.65$^{\scalebox{0.6}{±0.75}}$ \\

\hline
\multirow{4}{*}{\parbox{1.5cm}{\centering Places365}} 
& LT$\rightarrow$UNI & \xmark &  25.14$^{\scalebox{0.6}{±0.14}}$ &  17.45$^{\scalebox{0.6}{±0.30}}$ &  \textbf{27.77$^{\scalebox{0.6}{±0.45}}$} &           19.78$^{\scalebox{0.6}{±2.21}}$ &  \textbf{28.47$^{\scalebox{0.6}{±0.14}}$} &  32.99$^{\scalebox{0.6}{±0.46}}$ \\
& LT$\rightarrow$UNI & \cmark &  25.14$^{\scalebox{0.6}{±0.14}}$ &  16.24$^{\scalebox{0.6}{±1.39}}$ &  \textbf{27.69$^{\scalebox{0.6}{±0.51}}$} &           18.88$^{\scalebox{0.6}{±1.61}}$ &  \textbf{27.83$^{\scalebox{0.6}{±0.27}}$} &  33.38$^{\scalebox{0.6}{±0.31}}$ \\
& UNI$\rightarrow$LT & \xmark &  58.17$^{\scalebox{0.6}{±1.01}}$ &  81.16$^{\scalebox{0.6}{±0.61}}$ &  \textbf{81.64$^{\scalebox{0.6}{±0.63}}$} &  \textbf{82.04$^{\scalebox{0.6}{±0.15}}$} &           82.04$^{\scalebox{0.6}{±0.63}}$ &  88.14$^{\scalebox{0.6}{±0.27}}$ \\
& UNI$\rightarrow$LT & \cmark &  58.17$^{\scalebox{0.6}{±1.01}}$ &  81.20$^{\scalebox{0.6}{±0.61}}$ &  \textbf{81.65$^{\scalebox{0.6}{±0.61}}$} &           82.04$^{\scalebox{0.6}{±0.15}}$ &  \textbf{82.07$^{\scalebox{0.6}{±0.66}}$} &  88.15$^{\scalebox{0.6}{±0.30}}$ \\

\hline
\multirow{2}{*}{ImageNet}
& LT$\rightarrow$UNI & \xmark &  34.30$^{\scalebox{0.6}{±0.19}}$ &  19.02$^{\scalebox{0.6}{±0.26}}$ &  \textbf{33.57$^{\scalebox{0.6}{±0.33}}$} &           23.94$^{\scalebox{0.6}{±2.04}}$ &  \textbf{35.91$^{\scalebox{0.6}{±0.20}}$} &  37.34$^{\scalebox{0.6}{±0.15}}$ \\
& LT$\rightarrow$UNI & \cmark &  34.30$^{\scalebox{0.6}{±0.19}}$ &  17.28$^{\scalebox{0.6}{±0.48}}$ &  \textbf{32.34$^{\scalebox{0.6}{±0.41}}$} &           24.78$^{\scalebox{0.6}{±3.06}}$ &  \textbf{35.86$^{\scalebox{0.6}{±0.17}}$} &  37.39$^{\scalebox{0.6}{±0.16}}$ \\

\hline
\end{tabular}
\caption{\textit{``Improve Estimates from Confusion Matrix.'}' Accuracy (± std. dev.) after adaptation with new prior estimate based on confusion matrix (CM) inversion \cite{saerens2002adjusting} and our proposed method from Section \ref{subsubsection:cm_maximum_likelihood} (CM$^\text{L}$). SCM denotes soft confusion matrix, NA denotes no adaptation, Oracle is adaptation with ground truth priors. Results on CIFAR are averaged from 10 experiments, results on Places and ImageNet are averaged from 5 experiments. Best results are displayed in bold.}
\label{tab:cm_correction} 
\end{table*}

To evaluate the methods on practical tasks with prior shift, we experiment with fine-grained plant classification on the PlantCLEF data \cite{goeau2017plant,goeau2018overview} and with learning to classify ImageNet \cite{russakovsky2015imagenet} classes from a long-tailed noisy training dataset downloaded from the web, Webvision 1.0 \cite{li2017webvision}.

In the experiments with ImageNet and Webvision, we trained a ResNet-18 \cite{he2016deep} classifier with the standard input size 224x224 from scratch. In the experiments on Places365 and PlantCLEF, we finetuned ResNet-18 from an ImageNet-pretrained checkpoint. In the CIFAR100 experiments, we used a ResNet-32 adjusted to input size 32x32.

\subsection{New Priors Are Known: Adapt or Re-Train?}
\label{sec:exp_adapt_retrain}
Let us first examine the case when new class priors $p_\mathcal{E}$ are known, and compare:
\begin{enumerate}[noitemsep,topsep=0pt]
    \item Adapting the predictions of a previously trained classifier $f_\mathcal{T}(\mathbf{x}) \approx p_\mathcal{T}(Y|\mathbf x)$, following Eq. \eqref{eq:adapt_priors}.
    \item Training a classifier $f_\mathcal{E}(\mathbf{x})$ with a sampler following the known new class priors $p_\mathcal{E}$.
\end{enumerate}
When adapting predictions of classifier $f_\mathcal{T}$ following Eq. \eqref{eq:adapt_priors}, the trained priors can be determined either as a proportion of class labels in the training set, $\hat p^N_\mathcal{T}(Y=k) = \frac{N_k}{N}$, or as the average of predictions $f(\mathbf{x})$ on the training set, $\hat p^{f}_\mathcal{T}(Y) = \frac{1}{N}\sum\limits_{i=1}^n f(\mathbf{x_i}) $.

The training- and adaptation- strategies are experimentally compared on the CIFAR100-LT$^*$, Places365-LT and ImageNet-LT datasets in Table \ref{tab:adapt_or_retrain_cifar_pt}. The results show that adaptation of the classifier performs better than re-training the classifier with weighted sampling following $p_\mathcal{E}$. In most cases, the best results are achieved when trained priors are estimated from the predictions on the training set. We will thus estimate the trained priors by $\hat p_\mathcal{T}(Y) = \hat p^{f}_\mathcal{T}(Y)$.

\subsection{Prior Shift Estimation}

\subsubsection{Improving Estimates from Confusion Matrices}
Table \ref{tab:cm_correction} compares accuracy after adaptation with new prior estimate based on confusion matrix (CM) inversion \cite{saerens2002adjusting} and our proposed method from Section \ref{subsubsection:cm_maximum_likelihood} (CM$^\text{L}$) . The proposed method handles inconsistent estimates $\hat p(D)$ and $\mathbf{\hat{C}}_{d|y}$ and consistently improves the results both using the confusion matrix (CM$^\text{L}$) and the soft confusion matrix (SCM$^\text{L}$). In all cases, the proposed SCM$^\text{L}$ method using soft confusion matrix achieves the best results.

\begin{table*}[hbt] 
\centering
\setlength\tabcolsep{3pt}
\hspace*{-1.55mm}\begin{tabular}{m{1.6cm}|m{1.35cm}||>{\centering\arraybackslash}m{1.35cm}||>{\centering\arraybackslash}m{1.375cm}|>{\centering\arraybackslash}m{1.375cm}>{\centering\arraybackslash}m{1.375cm}>{\centering\arraybackslash}m{1.375cm}|>{\centering\arraybackslash}m{1.375cm}>{\centering\arraybackslash}m{1.375cm} >{\centering\arraybackslash}m{1.375cm}|>{\centering\arraybackslash}m{1.375cm}}
\multicolumn{2}{c||}{} & \multirow{2}{*}{\shortstack{BCTS \cite{alexandari2020maximum}\\ calibrated}} & \multirow{2}{*}{NA} & \multicolumn{3}{c|}{MLE} & \multicolumn{3}{c|}{MAP}  & \multirow{2}{*}{Oracle} \\
\multicolumn{2}{c||}{Dataset} &  &   & EM & CM$^\text{L}$ & SCM$^\text{L}$ & MAP & CM$^\text{M}$ & SCM$^\text{M}$ \\
\hline \hline
 \multirow{4}{*}{\parbox{1.6cm}{\centering CIFAR100$^*$}} 
& LT$\rightarrow$UNI & \xmark &  31.66$^{\scalebox{0.6}{±1.27}}$ &           32.81$^{\scalebox{0.6}{±1.41}}$ &  33.00$^{\scalebox{0.6}{±1.56}}$ &  \textbf{33.47$^{\scalebox{0.6}{±1.33}}$} &           32.73$^{\scalebox{0.6}{±1.42}}$ &           33.49$^{\scalebox{0.6}{±1.45}}$ &  \textbf{33.50$^{\scalebox{0.6}{±1.40}}$} &  34.06$^{\scalebox{0.6}{±1.35}}$ \\
& LT$\rightarrow$UNI & \cmark &  31.67$^{\scalebox{0.6}{±1.27}}$ &           29.43$^{\scalebox{0.6}{±1.59}}$ &  32.41$^{\scalebox{0.6}{±1.50}}$ &  \textbf{33.42$^{\scalebox{0.6}{±1.51}}$} &          24.46$^{\scalebox{0.6}{±12.43}}$ &           33.99$^{\scalebox{0.6}{±1.43}}$ &  \textbf{34.13$^{\scalebox{0.6}{±1.53}}$} &  34.40$^{\scalebox{0.6}{±1.38}}$ \\
& UNI$\rightarrow$LT & \xmark &  63.83$^{\scalebox{0.6}{±0.82}}$ &           67.23$^{\scalebox{0.6}{±0.88}}$ &  68.08$^{\scalebox{0.6}{±0.75}}$ &  \textbf{68.24$^{\scalebox{0.6}{±0.75}}$} &           66.72$^{\scalebox{0.6}{±0.91}}$ &  \textbf{67.01$^{\scalebox{0.6}{±0.91}}$} &           67.00$^{\scalebox{0.6}{±0.87}}$ &  69.13$^{\scalebox{0.6}{±0.58}}$ \\
& UNI$\rightarrow$LT & \cmark &  63.83$^{\scalebox{0.6}{±0.82}}$ &           69.17$^{\scalebox{0.6}{±0.91}}$ &  69.10$^{\scalebox{0.6}{±0.98}}$ &  \textbf{69.40$^{\scalebox{0.6}{±0.77}}$} &           68.30$^{\scalebox{0.6}{±0.77}}$ &  \textbf{68.42$^{\scalebox{0.6}{±0.84}}$} &           68.38$^{\scalebox{0.6}{±0.73}}$ &  70.65$^{\scalebox{0.6}{±0.75}}$ \\

\hline
\multirow{4}{*}{\parbox{1.5cm}{\centering Places365}} 
& LT$\rightarrow$UNI & \xmark &  25.14$^{\scalebox{0.6}{±0.14}}$ &           28.02$^{\scalebox{0.6}{±0.92}}$ &  27.77$^{\scalebox{0.6}{±0.45}}$ &  \textbf{28.47$^{\scalebox{0.6}{±0.14}}$} &           25.22$^{\scalebox{0.6}{±0.13}}$ &  \textbf{28.02$^{\scalebox{0.6}{±0.24}}$} &           27.68$^{\scalebox{0.6}{±0.13}}$ &  32.99$^{\scalebox{0.6}{±0.46}}$ \\
& LT$\rightarrow$UNI & \cmark &  25.14$^{\scalebox{0.6}{±0.14}}$ &  \textbf{28.09$^{\scalebox{0.6}{±1.32}}$} &  27.69$^{\scalebox{0.6}{±0.51}}$ &           27.83$^{\scalebox{0.6}{±0.27}}$ &  \textbf{28.57$^{\scalebox{0.6}{±0.24}}$} &           27.92$^{\scalebox{0.6}{±0.24}}$ &           27.41$^{\scalebox{0.6}{±0.15}}$ &  33.38$^{\scalebox{0.6}{±0.31}}$ \\
& UNI$\rightarrow$LT & \xmark &  58.17$^{\scalebox{0.6}{±1.01}}$ &  \textbf{82.63$^{\scalebox{0.6}{±0.31}}$} &  81.64$^{\scalebox{0.6}{±0.63}}$ &           82.04$^{\scalebox{0.6}{±0.63}}$ &  \textbf{76.97$^{\scalebox{0.6}{±0.45}}$} &           76.13$^{\scalebox{0.6}{±0.56}}$ &           73.27$^{\scalebox{0.6}{±0.46}}$ &  88.14$^{\scalebox{0.6}{±0.27}}$ \\
& UNI$\rightarrow$LT & \cmark &  58.17$^{\scalebox{0.6}{±1.01}}$ &  \textbf{82.63$^{\scalebox{0.6}{±0.26}}$} &  81.65$^{\scalebox{0.6}{±0.61}}$ &           82.07$^{\scalebox{0.6}{±0.66}}$ &  \textbf{77.00$^{\scalebox{0.6}{±0.41}}$} &           76.16$^{\scalebox{0.6}{±0.54}}$ &           73.30$^{\scalebox{0.6}{±0.46}}$ &  88.15$^{\scalebox{0.6}{±0.30}}$ \\

\hline
\multirow{2}{*}{ImageNet}
& LT$\rightarrow$UNI & \xmark &  34.30$^{\scalebox{0.6}{±0.19}}$ &           34.63$^{\scalebox{0.6}{±0.29}}$ &  33.57$^{\scalebox{0.6}{±0.33}}$ &  \textbf{35.91$^{\scalebox{0.6}{±0.20}}$} &           34.64$^{\scalebox{0.6}{±0.20}}$ &           36.41$^{\scalebox{0.6}{±0.17}}$ &  \textbf{36.57$^{\scalebox{0.6}{±0.16}}$} &  37.34$^{\scalebox{0.6}{±0.15}}$ \\
& LT$\rightarrow$UNI & \cmark &  34.30$^{\scalebox{0.6}{±0.19}}$ &           27.26$^{\scalebox{0.6}{±2.25}}$ &  32.34$^{\scalebox{0.6}{±0.41}}$ &  \textbf{35.86$^{\scalebox{0.6}{±0.17}}$} &          20.65$^{\scalebox{0.6}{±18.77}}$ &           36.18$^{\scalebox{0.6}{±0.12}}$ &  \textbf{36.80$^{\scalebox{0.6}{±0.14}}$} &  37.39$^{\scalebox{0.6}{±0.16}}$ \\

\hline
\end{tabular}
\caption{\textit{``How to estimate new priors?''} Accuracy (± std. dev.) after adaptation to new priors estimated with different Maximum Likelihood and Maximum A Posteriori estimates. NA denotes no adaptation, Oracle is adaptation with ground truth priors. Best MLE and MAP results are underlined for $f_\mathcal{T}$ and calibrated $f_\mathcal{T}$. Results on CIFAR are averaged from 10 experiments, results on Places and ImageNet are averaged from 5 experiments. Best results are displayed in bold.}
\label{tab:mle_map_cm_all} 
\end{table*}

\begin{table*}[h] 
\centering
\setlength\tabcolsep{3pt}
\hspace*{-1.55mm}\begin{tabular}{m{1.6cm}|m{1.35cm}||>{\centering\arraybackslash}m{1.35cm}||>{\centering\arraybackslash}m{1.90cm}|>{\centering\arraybackslash}m{1.90cm}>{\centering\arraybackslash}m{1.90cm}>{\centering\arraybackslash}m{1.90cm}>{\centering\arraybackslash}m{1.90cm}|>{\centering\arraybackslash}m{1.90cm}}
\multicolumn{2}{c||}{Dataset} & \shortstack{BCTS \cite{alexandari2020maximum}\\ calibrated} & NA & SCM$^\text{L}$ & RLLS & BBSE & BBSE-S & Oracle \\
\hline \hline
 \multirow{4}{*}{\parbox{1.6cm}{\centering CIFAR100$^*$}} 
& LT$\rightarrow$UNI & \xmark &  31.66$^{\scalebox{0.6}{±1.27}}$ &  \textbf{33.47$^{\scalebox{0.6}{±1.33}}$} &  32.75$^{\scalebox{0.6}{±1.40}}$ &  31.28$^{\scalebox{0.6}{±1.58}}$ &           31.92$^{\scalebox{0.6}{±1.62}}$ &  34.06$^{\scalebox{0.6}{±1.35}}$ \\
& LT$\rightarrow$UNI & \cmark &  31.67$^{\scalebox{0.6}{±1.27}}$ &  \textbf{33.42$^{\scalebox{0.6}{±1.51}}$} &  32.62$^{\scalebox{0.6}{±1.46}}$ &  26.47$^{\scalebox{0.6}{±1.87}}$ &           29.06$^{\scalebox{0.6}{±2.65}}$ &  34.40$^{\scalebox{0.6}{±1.38}}$ \\
& UNI$\rightarrow$LT & \xmark &  63.83$^{\scalebox{0.6}{±0.82}}$ &  \textbf{68.24$^{\scalebox{0.6}{±0.75}}$} &  68.02$^{\scalebox{0.6}{±0.77}}$ &  67.95$^{\scalebox{0.6}{±0.96}}$ &           68.12$^{\scalebox{0.6}{±0.90}}$ &  69.13$^{\scalebox{0.6}{±0.58}}$ \\
& UNI$\rightarrow$LT & \cmark &  63.83$^{\scalebox{0.6}{±0.82}}$ &           69.40$^{\scalebox{0.6}{±0.77}}$ &  69.05$^{\scalebox{0.6}{±0.97}}$ &  69.30$^{\scalebox{0.6}{±0.99}}$ &  \textbf{69.51$^{\scalebox{0.6}{±0.98}}$} &  70.65$^{\scalebox{0.6}{±0.75}}$ \\

\hline
\multirow{4}{*}{\parbox{1.5cm}{\centering Places365}} 
& LT$\rightarrow$UNI & \xmark &  25.14$^{\scalebox{0.6}{±0.14}}$ &  \textbf{28.47$^{\scalebox{0.6}{±0.14}}$} &  26.94$^{\scalebox{0.6}{±0.41}}$ &  24.79$^{\scalebox{0.6}{±0.74}}$ &           25.55$^{\scalebox{0.6}{±0.69}}$ &  32.99$^{\scalebox{0.6}{±0.46}}$ \\
& LT$\rightarrow$UNI & \cmark &  25.14$^{\scalebox{0.6}{±0.14}}$ &  \textbf{27.83$^{\scalebox{0.6}{±0.27}}$} &  27.03$^{\scalebox{0.6}{±0.40}}$ &  23.12$^{\scalebox{0.6}{±0.79}}$ &           23.68$^{\scalebox{0.6}{±0.75}}$ &  33.38$^{\scalebox{0.6}{±0.31}}$ \\
& UNI$\rightarrow$LT & \xmark &  58.17$^{\scalebox{0.6}{±1.01}}$ &  \textbf{82.04$^{\scalebox{0.6}{±0.63}}$} &  82.04$^{\scalebox{0.6}{±0.69}}$ &  80.66$^{\scalebox{0.6}{±0.57}}$ &           81.69$^{\scalebox{0.6}{±0.20}}$ &  88.14$^{\scalebox{0.6}{±0.27}}$ \\
& UNI$\rightarrow$LT & \cmark &  58.17$^{\scalebox{0.6}{±1.01}}$ &  \textbf{82.07$^{\scalebox{0.6}{±0.66}}$} &  82.04$^{\scalebox{0.6}{±0.69}}$ &  80.71$^{\scalebox{0.6}{±0.56}}$ &           81.69$^{\scalebox{0.6}{±0.20}}$ &  88.15$^{\scalebox{0.6}{±0.30}}$ \\

\hline
\multirow{2}{*}{ImageNet}
& LT$\rightarrow$UNI & \xmark &  34.30$^{\scalebox{0.6}{±0.19}}$ &  \textbf{35.91$^{\scalebox{0.6}{±0.20}}$} &  34.69$^{\scalebox{0.6}{±0.14}}$ &  30.77$^{\scalebox{0.6}{±0.31}}$ &           31.31$^{\scalebox{0.6}{±0.90}}$ &  37.34$^{\scalebox{0.6}{±0.15}}$ \\
& LT$\rightarrow$UNI & \cmark &  34.30$^{\scalebox{0.6}{±0.19}}$ &  \textbf{35.86$^{\scalebox{0.6}{±0.17}}$} &  34.31$^{\scalebox{0.6}{±0.08}}$ &  26.89$^{\scalebox{0.6}{±0.49}}$ &           28.05$^{\scalebox{0.6}{±1.69}}$ &  37.39$^{\scalebox{0.6}{±0.16}}$ \\

\hline
\end{tabular}
\caption{\text{``Estimate test priors or directly the prior ratio?''} Accuracy (± std. dev.) after adaptation with the priors estimated by SCM$^\text{L}$ or with the prior ratio estimated by BBSE \cite{lipton2018detecting} and RLLS \cite{azizzadenesheli2019regularized} (without re-training). Results on CIFAR are averaged from 10 experiments, results on Places and ImageNet are averaged from 5 experiments. Best results are displayed in bold.}
\label{tab:prior_ratio_noretrain} 
\end{table*}

\subsubsection{Methods for MLE and MAP Prior Estimation}

Existing methods for maximum likelihood and maximum a-posteriori prior estimation are compared against the methods proposed in Sections \ref{subsubsection:cm_maximum_likelihood} and \ref{subsubsection:cm_maximum_aposteriori} respectively in Table \ref{tab:mle_map_cm_all}. 
Note that the methods maximize a different likelihood function: The EM algorithm of Saerens et al. \cite{saerens2002adjusting} maximizes the likelihood of observed classifier outputs $f(\mathbf{x}_i)$, while the proposed methods based on confusion matrix (CM$^L$) and soft confusion matrix  (SCM$^L$) maximize the likelihood of classifiers decisions $\argmax_k f(\mathbf{x}_i)$. The same difference in likelihood functions holds for the MAP approach of Sulc and Matas \cite{sulc2019improving} and MAP estimate proposed in Section \ref{subsubsection:cm_maximum_aposteriori}, but we use the same hyper-prior on $p_\mathcal{E}(Y)$ for all methods: $Dir(\alpha=3)$.

From the maximum likelihood estimators, the proposed SCM$^\text{L}$ achieves the best results in most cases, with the exception of Places365 "UNI$\rightarrow$LT", where the EM algorithm performed slightly better. 
Similarly, the Maximum A-Posteriori version of the proposed method, SCM$^\text{M}$ performs better than the existing MAP estimate \cite{sulc2019improving} in most cases. As expected, the MAP estimation improves upon MLE on the dense test distributions, favoured by the Dirichlet hyper-prior.

\subsection{Prior Ratio Estimation}
Having an estimate of the trained priors, Table \ref{tab:prior_ratio_noretrain} compares prior ratio estimation with BBSE, BBSE-S \cite{lipton2018detecting} and RLLS \cite{azizzadenesheli2019regularized} against the best performing prior estimation methods, CM$^\text{L}$ and SCM$^\text{L}$. The results indicate that it is better to estimate the new priors than to directly estimate the prior ratio with BBSE or BBSE-S.

\subsection{Dependence on the Number of Test Samples}
\begin{figure}[htb]
\centering
\hspace*{-0.15cm}
\includegraphics[width=.48\textwidth]{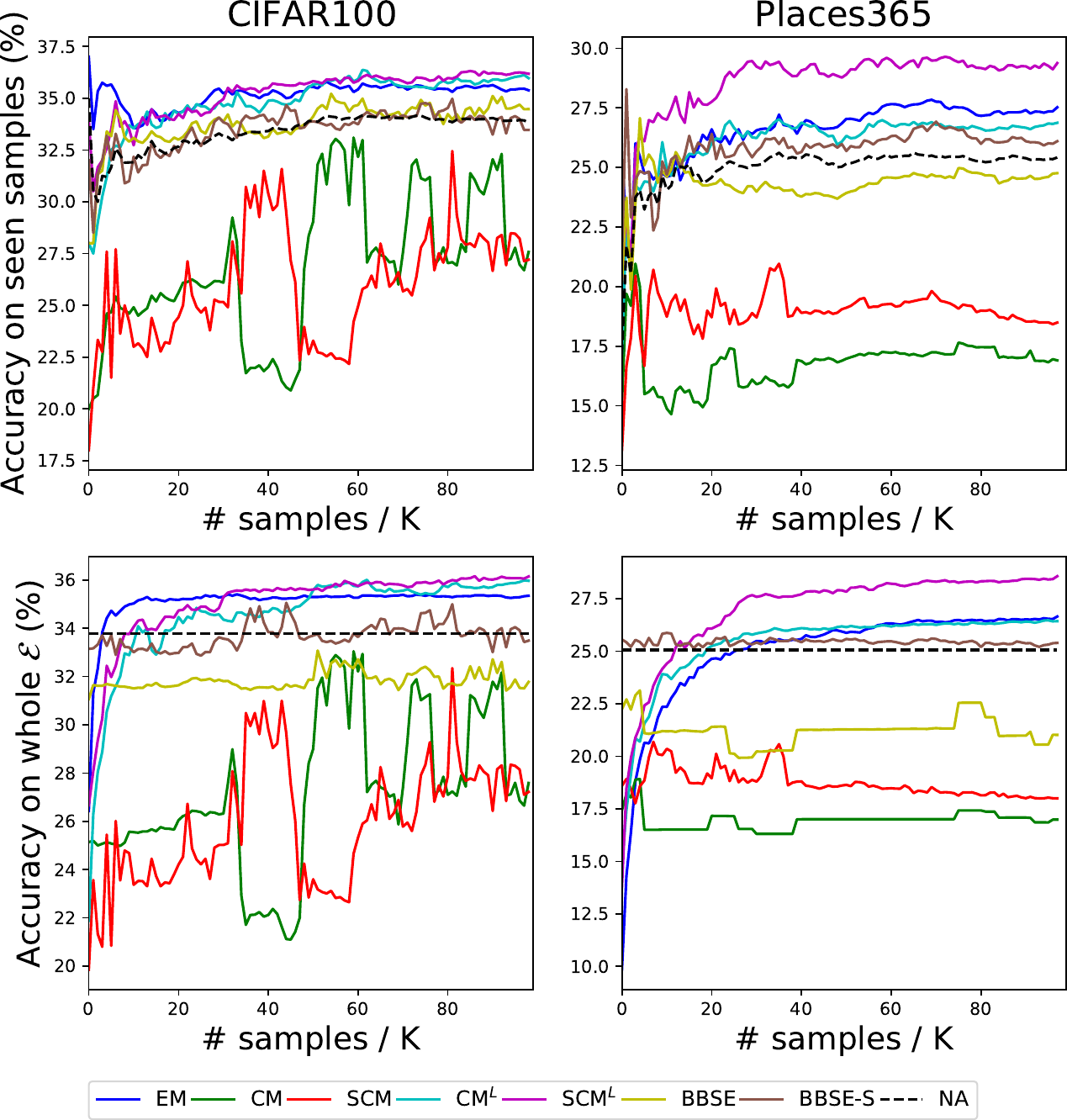}
\caption{\textit{``How many samples do I need?''} Accuracy after adapting CIFAR100-LT$\rightarrow$UNI (left) and Places365-LT$\rightarrow$UNI (right) using \#samples for prior estimation.}
\label{fig:test_num_samples}
\end{figure}

Figure \ref{fig:test_num_samples} displays the accuracy on the uniformly distributed sets after adaptation of classifiers trained on the CIFAR100-LT$^*$ and Places365-LT datasets with different prior estimation methods, as a function of the number of test examples used for prior estimation. While the proposed SCM$^\text{L}$ method achieves slightly higher accuracy with more samples, the EM algorithm works slightly better with low number of samples. With extremely low number of samples, prior estimation should be omitted.

\subsection{Applying the Best Practice}
The lessons learned were applied to two tasks with naturally imbalanced priors:

A classifier trained on the imbalanced Webvision 1.0 dataset achieved 57.12\% accuracy on the ImageNet validation set. ImageNet has a uniform class distribution. We calibrated the classifier with BCTS \cite{alexandari2020maximum}, estimated the trained prior, and adapted the predictions by Equation \eqref{eq:adapt_priors}. The classification accuracy improved to 58.22\%.

A classifier of 10,000 plant species trained on the PlantCLEF 2017 dataset (EOL+test) achieved 37.95\% accuracy on the PlantCLEF 2018 test set of 2072 images. Because the number of samples was very low, we used the EM algorithm \cite{alexandari2020maximum,saerens2002adjusting} on the calibrated predictions. Adapting to the estimated priors increased the accuracy to 41.35\%.

\section{Conclusions}
This paper reviews and compares existing methods for adaptation to prior shift and proposes a novel method to deal with a known problem of existing methods based on confusion matrices \cite{forman2008quantifying,saerens2002adjusting,vucetic2001classification}, where inconsistent
estimates of decision probabilities and confusion matrices can result in negative values in the estimated priors. The proposed method, (S)CM$^\text{L}$, deals with this problem by constrained maximization of the likelihood of classifier decisions on the new test set. It has further been extended into a Maximum A-Posteriori estimator (S)CM$^\text{M}$ by adding a hyper-prior on the new prior distribution.

Experimental analysis of the existing and proposed methods for prior shift adaptation suggests the following best practice:
\begin{itemize}[noitemsep,topsep=0pt,leftmargin=*]
    \item Adaptation of the original classifier typically performs better than re-training
the classifier with sampling matching the shift, and is significantly computationally cheaper.
    \item The proposed method handles inconsistent estimates $\hat p(D)$ and $\mathbf{\hat{C}}_{d|y}$ and consistently improves the results both using the confusion matrix (CM$^\text{L}$) and the soft confusion matrix (SCM$^\text{L}$).
    \item From the compared maximum likelihood estimators, the proposed SCM$^\text{L}$ achieves the best results in most cases. 
    \item The EM algorithm \cite{saerens2002adjusting,alexandari2020maximum} works better with a low number of samples. With extremely low number of samples, prior estimation should better be omitted at all.
    \item The proposed Maximum A-Posteriori approach, SCM$^\text{M}$, performs better than the existing MAP estimate \cite{sulc2019improving}.
    \item If trained priors can be estimated, it is better to estimate the test set priors than to directly estimate the prior ratio with BBSE or RLLS.
    \item Prior shift adaptation relies on a well-calibrated classifier, assumed in Eq. \eqref{eq:adapt_priors}. In \cite{alexandari2020maximum}, BCTS improves prior shift adaptation of classifiers trained on uniform distribution, similarly to our UNI$\rightarrow$LT experiments. With class-specific parameters, BCTS may overfit to errors on the validation set. For classifiers trained on LT datasets, we show BCTS is not a reliable calibration method, as it often decreases the final recognition accuracy.
\end{itemize}
Applying the best practice to two tasks with naturally imbalanced priors, learning from web-crawled images and plant classification, increased the accuracy by 1.1\% and 3.4\% respectively.

\paragraph{Acknowledgment}
This work was supported by the CTU student grant SGS20/171/OHK3/3T/13, by Toyota Motor Europe, and by the OP VVV project CZ.02.1.01/0.0/0.0/16 019/0000765
Research Center for Informatics.

{\small
\bibliographystyle{ieee_fullname}
\bibliography{bibliography}

\begin{thebibliography}{10}\itemsep=-1pt

\bibitem{alexandari2020maximum}
Amr Alexandari, Anshul Kundaje, and Avanti Shrikumar.
\newblock Maximum likelihood with bias-corrected calibration is hard-to-beat at
  label shift adaptation.
\newblock In {\em ICML}, pages 222--232, 2020.

\bibitem{azizzadenesheli2019regularized}
Kamyar Azizzadenesheli, Anqi Liu, Fanny Yang, and Animashree Anandkumar.
\newblock Regularized learning for domain adaptation under label shifts.
\newblock In {\em ICLR}, 2019.

\bibitem{cao2019learning}
Kaidi Cao et~al.
\newblock Learning imbalanced datasets with label-distribution-aware margin
  loss.
\newblock In {\em NeurIPS}, 2019.

\bibitem{cui2018large}
Yin Cui, Yang Song, Chen Sun, Andrew Howard, and Serge Belongie.
\newblock Large scale fine-grained categorization and domain-specific transfer
  learning.
\newblock In {\em Proceedings of the IEEE conference on computer vision and
  pattern recognition}, pages 4109--4118, 2018.

\bibitem{duplessis2012semi}
Marthinus~Christoffel du Plessis and Masashi Sugiyama.
\newblock Semi-supervised learning of class balance under class-prior change by
  distribution matching.
\newblock {\em CoRR}, abs/1206.4677, 2012.

\bibitem{forman2008quantifying}
George Forman.
\newblock Quantifying counts and costs via classification.
\newblock {\em Data Mining and Knowledge Discovery}, 17(2):164--206, Oct 2008.

\bibitem{goeau2017plant}
Herve Goeau, Pierre Bonnet, and Alexis Joly.
\newblock Plant identification based on noisy web data: the amazing performance
  of deep learning (lifeclef 2017).
\newblock 2017.

\bibitem{goeau2018overview}
Herv{\'e} Go{\"e}au, Pierre Bonnet, and Alexis Joly.
\newblock Overview of expertlifeclef 2018: how far automated identification
  systems are from the best experts?
\newblock 2018.

\bibitem{guo2017calibration}
Chuan Guo, Geoff Pleiss, Yu Sun, and Kilian~Q Weinberger.
\newblock On calibration of modern neural networks.
\newblock In {\em ICML}, pages 1321--1330, 2017.

\bibitem{he2016deep}
Kaiming He, Xiangyu Zhang, Shaoqing Ren, and Jian Sun.
\newblock Deep residual learning for image recognition.
\newblock In {\em Proceedings of the IEEE conference on computer vision and
  pattern recognition}, pages 770--778, 2016.

\bibitem{Idelbayev18a}
Yerlan Idelbayev.
\newblock Proper {ResNet} implementation for {CIFAR10/CIFAR100} in {PyTorch}.
\newblock \url{https://github.com/akamaster/pytorch_resnet_cifar10}.
\newblock Accessed: 2021-03-23.

\bibitem{kozerawski2020blt}
Jedrzej Kozerawski, Victor Fragoso, Nikolaos Karianakis, Gaurav Mittal, Matthew
  Turk, and Mei Chen.
\newblock Blt: Balancing long-tailed datasets with adversarially-perturbed
  images.
\newblock In {\em ACCV}, 2020.

\bibitem{krizhevsky2009learning}
Alex Krizhevsky.
\newblock Learning multiple layers of features from tiny images.
\newblock Technical report, 2009.

\bibitem{li2017webvision}
Wen Li, Limin Wang, Wei Li, Eirikur Agustsson, and Luc Van~Gool.
\newblock Webvision database: Visual learning and understanding from web data.
\newblock {\em arXiv preprint arXiv:1708.02862}, 2017.

\bibitem{lin2017focal}
Tsung-Yi Lin et~al.
\newblock Focal loss for dense object detection.
\newblock In {\em CVPR}, 2017.

\bibitem{lipton2018detecting}
Zachary~C. Lipton, Yu-Xiang Wang, and Alex Smola.
\newblock Detecting and correcting for label shift with black box predictors,
  2018.

\bibitem{liu2019large}
Ziwei Liu et~al.
\newblock Large-scale long-tailed recognition in an open world.
\newblock In {\em CVPR\textbf{}}, 2019.

\bibitem{mclachlan1992discriminant}
Geoffrey~J. McLachlan.
\newblock {\em Discriminant Analysis and Statistical Pattern Recognition}.
\newblock John Wiley \& Sons, Inc, Hoboken, NJ, USA, 1992-03-27.

\bibitem{russakovsky2015imagenet}
Olga Russakovsky, Jia Deng, Hao Su, Jonathan Krause, Sanjeev Satheesh, Sean Ma,
  Zhiheng Huang, Andrej Karpathy, Aditya Khosla, Michael Bernstein, et~al.
\newblock Imagenet large scale visual recognition challenge.
\newblock {\em International journal of computer vision}, 115(3):211--252,
  2015.

\bibitem{saerens2002adjusting}
Marco Saerens, Patrice Latinne, and Christine Decaestecker.
\newblock Adjusting the outputs of a classifier to new a priori probabilities:
  A simple procedure.
\newblock {\em Neural Comput.}, 14(1):21–41, Jan. 2002.

\bibitem{sugiyama2008direct}
Masashi Sugiyama et~al.
\newblock Direct importance estimation with model selection and its application
  to covariate shift adaptation.
\newblock In {\em NIPS}, 2008.

\bibitem{sulc2019improving}
Milan Sulc and Jiri Matas.
\newblock Improving cnn classifiers by estimating test-time priors.
\newblock In {\em Proceedings of the IEEE/CVF International Conference on
  Computer Vision (ICCV) Workshops}, Oct 2019.

\bibitem{vucetic2001classification}
Slobodan Vucetic and Zoran Obradovic.
\newblock Classification on data with biased class distribution.
\newblock In {\em European Conference on Machine Learning}, pages 527--538.
  Springer, 2001.

\bibitem{wang2013projection}
Weiran Wang and Miguel~{\'A}. Carreira-Perpi{\~n}{\'a}n.
\newblock Projection onto the probability simplex: An efficient algorithm with
  a simple proof, and an application.
\newblock {\em ArXiv}, abs/1309.1541, 2013.

\bibitem{zadrozny2004learning}
Bianca Zadrozny.
\newblock Learning and evaluating classifiers under sample selection bias.
\newblock In {\em ICML}, 2004.

\bibitem{zhang2013domain}
Kun Zhang et~al.
\newblock Domain adaptation under target and conditional shift.
\newblock In {\em ICML}, 2013.

\bibitem{zhou2014learning}
Bolei Zhou, Agata Lapedriza, Jianxiong Xiao, Antonio Torralba, and Aude Oliva.
\newblock Learning deep features for scene recognition using places database.
\newblock In {\em Advances in neural information processing systems}, pages
  487--495, 2014.

\end{thebibliography}
}

\onecolumn

\title{Supplementary material to\\The Hitchhiker’s Guide to Prior-Shift Adaptation}
\author{}
\date{\vspace{-10ex}}
\maketitle

\renewcommand\thesection{\Alph{section}}
\setcounter{section}{0}

\section{Detail Description of the Experimental Setting} \label{sec:experimental_setting}
In the experiments conducted on ImageNet and Places365, we used the ResNet-18 \cite{he2016deep} classifier architecture. On ImageNet, the networks were trained from scratch for 90 epochs with the Stochastic Gradient Descent (SGD), initial learning rate set to 0.1 and decaying every 30 epochs by a factor of 10. On Places365, the network was finetuned from an ImageNet-pretrained checkpoint for 30 epochs with the initial learning rate set to 0.01 and decaying by a factor of 10 every 10 epochs.  For experiments on CIFAR100, we trained a ResNet-32 \cite{Idelbayev18a} adjusted to image input of 32x32. The network was trained from scratch for a 200 epochs, with a learning rate of 0.1 and decaying every 80 epochs by a factor of 10.  In all experiments, the batch size was set to 256. Momentum and weight decay were set to 0.9 and 0.0001 respectively.

The proposed methods (S)CM$^L$ takes two hyperparameter - learning rate and the number of steps in gradient ascent. Since the method has a well-defined objectives, we can find the learning rate by solving the task for several different rates and then select the one maximizing objective given by the Equation \eqref{eq:cm_max_problem}. The number of steps is set to 1000 in all experiments. The (S)CM$^M$ has three hyper-parameters: learning rate, number of steps, and $\alpha$ - the parameter of the symmetric Dirichlet distribution. We search only for the learning rate by maximizing objective given by the Equation \eqref{maximum_aposteriori_CM} and set $\alpha = 3$, following \cite{sulc2019improving}, and the number of steps again to 1000. The results for RLLS are based on the authors' code\footnote{\url{https://github.com/Angie-Liu/labelshift/tree/5bbe517938f4e3f5bd14c2c105de973dcc2e0917}}, using the \textit{cvxpy} optimizer. As termination condition for the EM and MAP algorithms, we set a threshold to $l_2$-distance between two consecutive solutions. The threshold is set to 0.001.

\section{Experimental Results on Additional Prior Distributions}

\begin{figure*}[bht]
\centering
\hspace*{-0.4cm}
\includegraphics[width=1.05\textwidth]{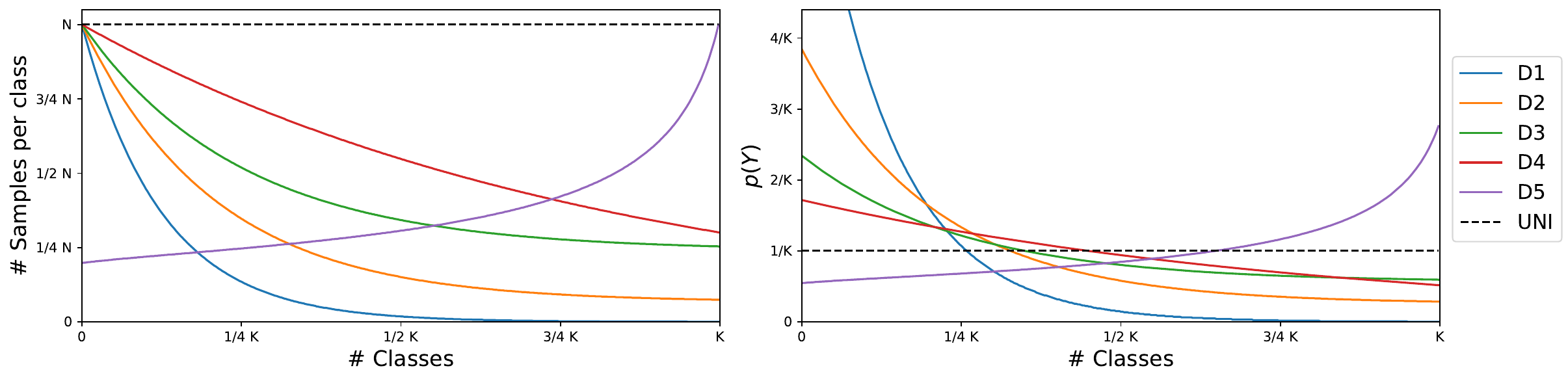}
\caption{Long-tailed class distributions D1, D2, D3, D4, D5 used to sample experimental subsets of CIFAR-100, Places-365.}
\label{fig:lt_distribution_sup}
\end{figure*}

In order to experiment with different prior shifts, we additionally sub-sampled several subsets of CIFAR-100 \cite{krizhevsky2009learning} and Places-365 \cite{zhou2014learning}, following different imbalanced categorical distributions displayed in Figure \ref{fig:lt_distribution_sup}. While all the proposed distributions are used for evaluation, only the original uniform distribution (UNI) and the distribution denoted as D3 were used for training\footnote{In order to reduce the computational load.}.

The results are displayed in Tables \ref{tab:cm_correction_sup}, \ref{tab:mle_map_cm_all_sup}, \ref{tab:prior_ratio_noretrain_sup} which are formatted analogically to Tables 2, 3, 4 in the paper manuscript.

\section{Inconsistent Estimates in Prior Ratio Estimation}
Similarly to the (S)CM methods  \cite{mclachlan1992discriminant,saerens2002adjusting}, the prior ratio estimation in BBSE \cite{lipton2018detecting} also suffers from the problem of inconsistent estimates. Let us consider the following example, resulting in a negative value in the estimated prior ratio, which -- as a ratio of two probabilities -- should be non-negative.

\begin{equation}
  \mathbf{\hat{C}}_{d,y}=\begin{bmatrix}0.4 & 0.1\\ 0.1 & 0.4\end{bmatrix}, \; \hat p(D) = \begin{bmatrix}1 \\ 0\end{bmatrix} \rightarrow \hat w(Y) = \begin{bmatrix} \frac{8}{3} \\ -\frac{2}{3}\end{bmatrix}   
\label{eq:bbse_inconsistent}
\end{equation}

\section{Confusion Matrices Illustrated on an Artificial Dataset}
Let us consider an illustrative classification problem with two classes $\{0,1\}$, generated from known normal distributions: $p(x|Y=0)=\mathcal{N}(-2,2)$ and $p(x|Y=1)=\mathcal{N}(2,2)$ with equal priors, $p(Y=0)=p(Y=1)=0.5$. We use 3 different classifiers modeled by logistic regression in the form: 
\begin{equation}
f(x)=\frac{1}{1+e^{-(ax+b)}}    
\end{equation}

\begin{figure}[b]
    \centering
	\includegraphics[width=1.0\linewidth]{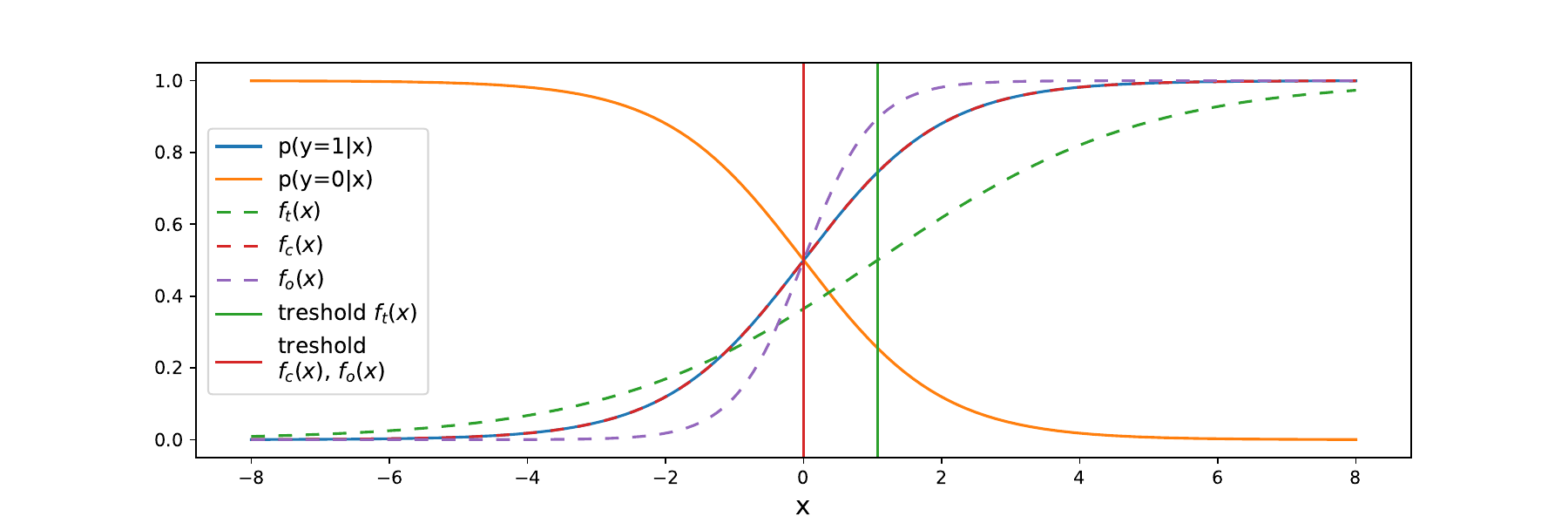}
	\caption{An illustrative 2-class example with known class posteriors (solid curves), outputs of 3 classifiers 
	for given $x$ (dashed lines) and their decision thresholds (vertical lines). Note that $f_{c}(x)$ and $f_{o}(x)$ are optimal Bayes classifiers minimizing the 0/1 loss.}
	\label{fig:artificial_ex_classif}
\end{figure}

The first classifier, $f_t(x)$, was trained with \textit{scikit-learn} on 4 randomly generated samples. The other two are Bayes classifiers: $f_{c}(x)$ is perfectly calibrated with parameters $a=1$ and $b=0$; $f_{o}(x)$ is overconfident with parameters $a=2$ and $b=0$. All three classifiers are illustrated together with their decision thresholds and the known posterior probabilities in Figure \ref{fig:artificial_ex_classif}.

Following Section \ref{subsection:cm_prior_estimation}, we denote $\mathbf{\hat C}_{d|y}$ the confusion matrix estimated from top-1 predictions and $\mathbf{\hat C}^{soft}_{d|y}$ the confusion matrix estimated using softmax outputs, following Eq. \eqref{eq:soft_estimate}. In this artificial example with known distributions, we can compute the \textit{true confusion matrix} $\mathbf{C}_{d|y}$ as follows:

\begin{equation}\label{eq:hard_bin_classif}
    \mathbf{C}_{1|j}=\int_t^{\infty}p(x|Y=j)\, dx; \quad \mathbf{C}_{0|j}=\int^t_{-\infty}p(x|Y=j)\,dx
\end{equation}

\begin{figure}
    \centering
    
         \begin{subfigure}[b]{0.31\textwidth}
         \centering
         \caption{Classifier $f_t(x)$ trained on 4 samples:}
\includegraphics[width=1.0\linewidth]{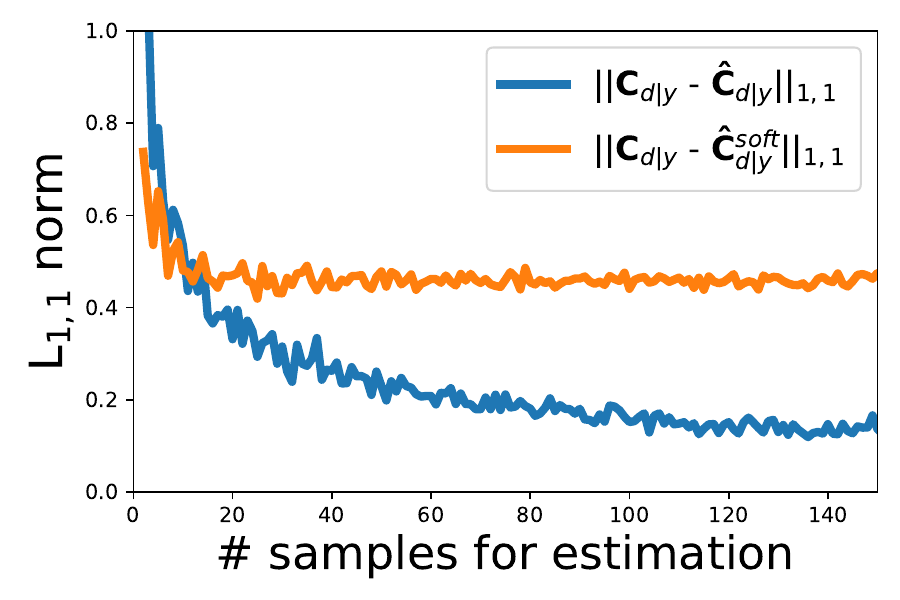}\\
{\hspace*{-8mm} $ \setlength\arraycolsep{2pt}
\scriptstyle \underset{\mathbf{\hat C}_{d|y}}{ \begin{bmatrix} 0.94 & 0.30 \\ 0.06 & 0.70 \end{bmatrix}}, \underset{\mathbf{\hat C}^{soft}_{d|y}}{\begin{bmatrix} 0.785 & 0.396 \\ 0.215 & 0.604 \end{bmatrix}}, \underset{\mathbf{C}_{d|y}}{ \begin{bmatrix} 0.938 & 0.321 \\ 0.062 & 0.679 \end{bmatrix}}$}
        \end{subfigure}
        \hfill
        \begin{subfigure}[b]{0.31\textwidth}
         \centering
         \caption{Overconfident classifier $f_{o}(x)$:}
\includegraphics[width=1.0\linewidth]{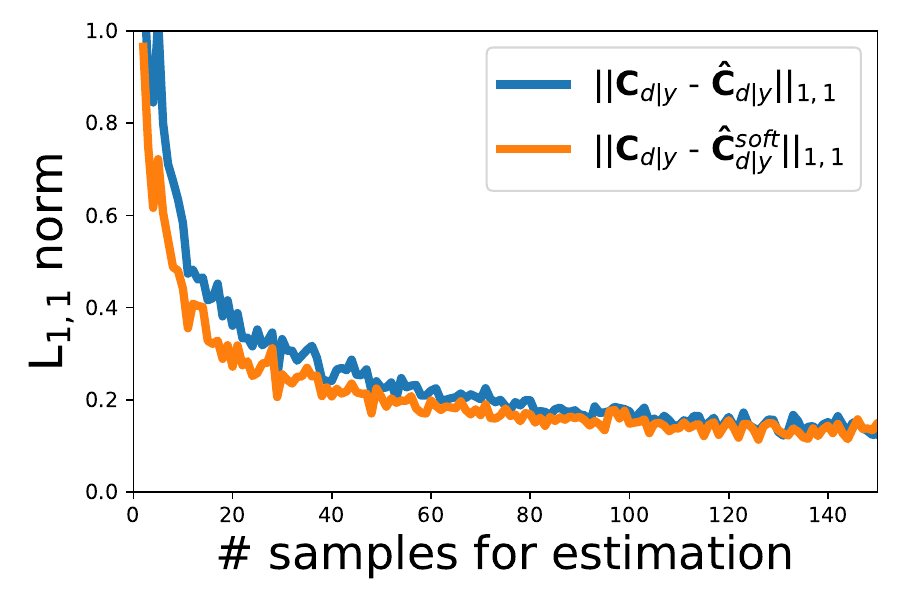}\\
{\hspace*{-4mm}
$ \setlength\arraycolsep{2pt}
\scriptstyle \underset{\mathbf{\hat C}_{d|y}}{ \begin{bmatrix} 0.82 & 0.16 \\ 0.18 & 0.84 \end{bmatrix}}, \underset{\mathbf{\hat C}^{soft}_{d|y}}{\begin{bmatrix} 0.798 & 0.174 \\ 0.202 & 0.826 \end{bmatrix}}, \underset{\mathbf{C}_{d|y}}{ \begin{bmatrix} 0.841 & 0.159 \\ 0.159 & 0.841 \end{bmatrix}}$}
        \end{subfigure}
        \hfill
        \begin{subfigure}[b]{0.31\textwidth}
         \centering
\caption{Calibrated classifier $f_{c}(x)$:}
\includegraphics[width=1.0\linewidth]{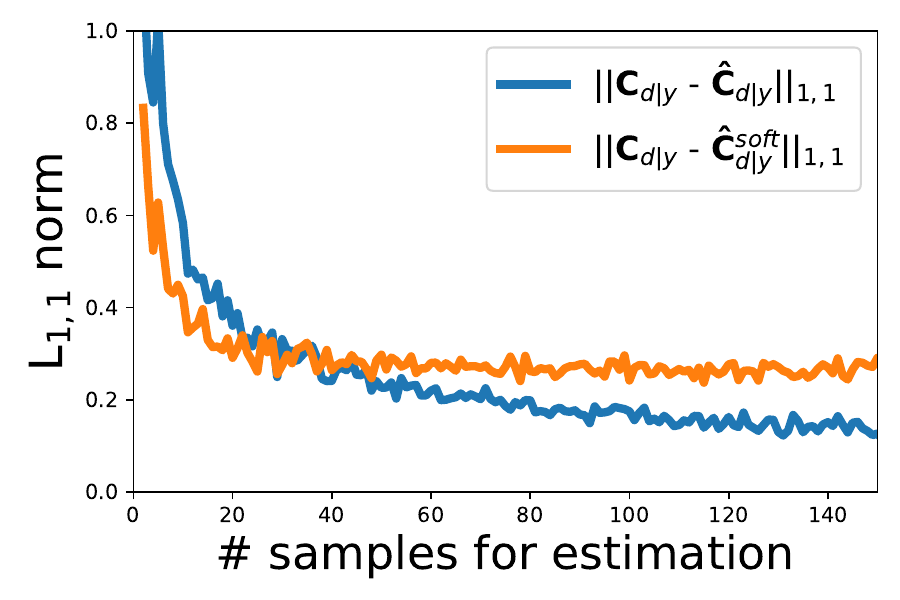}\\
{\hspace*{1mm}$ \setlength\arraycolsep{2pt}
\scriptstyle \underset{\mathbf{\hat C}_{d|y}}{ \begin{bmatrix} 0.82 & 0.16 \\ 0.18 & 0.84 \end{bmatrix}}, \underset{\mathbf{\hat C}^{soft}_{d|y}}{\begin{bmatrix} 0.764 & 0.215 \\ 0.236 & 0.785 \end{bmatrix}}, \underset{\mathbf{C}_{d|y}}{ \begin{bmatrix} 0.841 & 0.159 \\ 0.159 & 0.841 \end{bmatrix}}$ }\\
        \end{subfigure}
        
	\caption{Top: The distance (sum of absolute differences) of estimated confusion matrix $\mathbf{\hat C}_{d|y}$ and soft confusion matrix $\mathbf{\hat C}^{soft}_{d|y}$ from the true confusion matrix $\mathbf{C}_{d|y}$, depending on the number of validation samples. The distance values are averaged over 50 trials. Bottom: Confusion matrices $\mathbf{\hat C}_{d|y}$, $\mathbf{\hat C}^{soft}_{d|y}$ estimated from 50 samples and the true confusion matrix $\mathbf{C}_{d|y}$ computed from Equation \eqref{eq:hard_bin_classif}.}
	\label{fig:diff_learned}
\end{figure}

Figure \ref{fig:diff_learned} compares the distance of the two estimated confusion matrices from the true confusion matrix, depending on the size of the validation set. Note that the soft confusion matrix may not converge to the true confusion matrix even with a perfectly calibrated classifier, but it provides a better estimate in low-sample scenarios.

\section{Convergence Speed}
\begin{figure}[h]
    \centering
	\includegraphics[width=0.6\linewidth]{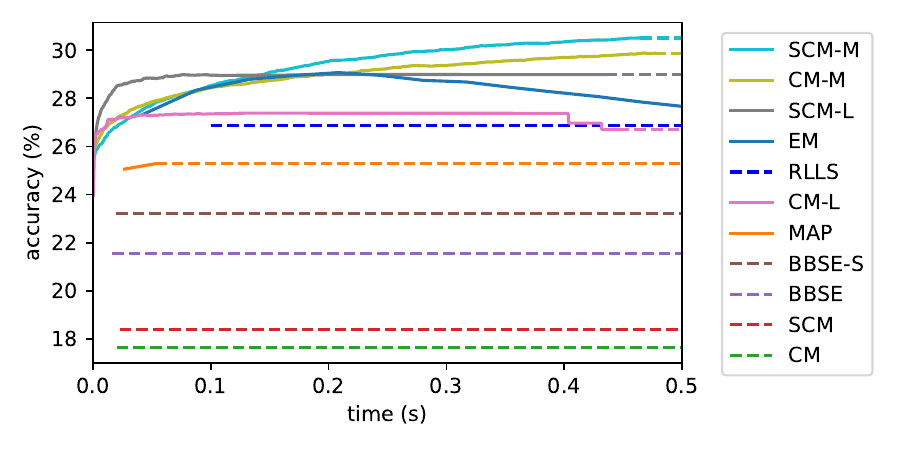}
	\caption{Convergence of prior estimation on Places365-LT. The \textit{y}-axis shows the classification accuracy after prior shift adaptation. Bold lines depict a particular method before termination criterion is met. Dashed lines show the constant accuracy after the computation finished. Note (S)CM, BBSE and BBSE-S are single-step methods. Since RLLS uses a third party optimizer, we only report the the total optimization time.}
	\label{fig:runtimes}
\end{figure}

The convergence of all algorithms on the test set of Places365-LT is displayed in Figure \ref{fig:runtimes}. Note that the optimization code for our methods is experimental and not optimized for run time. The results for RLLS are based on the authors' code\footnote{\url{https://github.com/Angie-Liu/labelshift/tree/5bbe517938f4e3f5bd14c2c105de973dcc2e0917}}, using the \textit{cvxpy} optimizer.
While the proposed optimization in (S)CM$^L$ and (S)CM$^M$ takes longer than the baseline (S)CM, all methods converge within 0.5 seconds on Places365-LT. Table \ref{tab:runtimes} compares the runtimes of all algorithms on Cifar100-LT, Places365-LT and ImageNet-LT, using the termination conditions used in all our experiments and described in Section \ref{sec:experimental_setting}. While the proposed optimization in (S)CM$^L$ and (S)CM$^M$ takes longer than the baseline (S)CM, all methods converge within 0.5 seconds on Places365-LT. Even though the predictions are adapted on CPU and the classifier evaluation is computed on GPU, the time to adapt classifier to label shift is negligible compared to the time it takes to evaluate the classifier on the test set, which takes several minutes on ImageNet-LT.

\begin{table}[h]
    \centering
    \begin{tabular}{c|ccc}
         & Cifar100-LT & Places365-LT & ImageNet-LT \\
        \hline
        CM      & 0.0038 & 0.0697 & 0.3486 \\
        SCM     & 0.0034 & 0.0555 & 0.2946 \\
        BBSE    & 0.0089 & 0.0678 & 0.3357 \\
        BBSE-S  & 0.0033 & 0.0506 & 0.2927 \\
        EM      & 0.0068 & 0.8293 & 2.2329 \\
        CM-L    & 0.2001 & 0.3104 & 1.3035 \\
        SCM-L   & 0.1828 & 0.2955 & 1.3595 \\
        MAP     & 0.0041 & 0.0916 & 0.3388 \\
        CM-M    & 0.2061 & 0.3355 & 1.3943 \\
        SCM-M   & 0.1998 & 0.3114 & 1.3510 \\
        RLLS    & 0.0268 & 0.1946 & 2.1280 \\

    \end{tabular}
    \caption{Run time (in seconds) of methods estimating test time prior $p_{\mathcal{E}}(Y)$ or prior ratio $p_{\mathcal{E}}(Y)/p_{\mathcal{T}}(Y)$. The run time was measured on a laptop with Intel® Core™ i7-6700HQ CPU @ 2.60GHz × 8 and averaged over 10 runs.}
    \label{tab:runtimes}
\end{table}

\begin{table*}[hbt] 
\centering
\setlength\tabcolsep{3pt}
\hspace*{-1.55mm}\begin{tabular}{>{\centering\arraybackslash}m{1.cm}|m{1.6cm}||>{\centering\arraybackslash}m{1.35cm}||>{\centering\arraybackslash}m{1.90cm}|>{\centering\arraybackslash}m{1.90cm}>{\centering\arraybackslash}m{1.90cm}|>{\centering\arraybackslash}m{1.90cm}>{\centering\arraybackslash}m{1.90cm}|>{\centering\arraybackslash}m{1.90cm}}
\multicolumn{2}{c||}{Dataset} & \shortstack{BCTS \cite{alexandari2020maximum}\\ calibrated} & NA & CM  & CM$^\text{L}$  & SCM  & SCM$^\text{L}$ & Oracle \\
\hline \hline
 \multirow{24}{*}{ \rotatebox[origin=c]{90}{CIFAR-100}} 
& D3$\rightarrow$D1 & \xmark &  62.76$^{\scalebox{0.6}{±1.60}}$ &           65.45$^{\scalebox{0.6}{±2.54}}$ &  \textbf{66.94$^{\scalebox{0.6}{±1.66}}$} &           66.45$^{\scalebox{0.6}{±2.04}}$ &  \textbf{67.35$^{\scalebox{0.6}{±1.56}}$} &  69.18$^{\scalebox{0.6}{±1.36}}$ \\
& D3$\rightarrow$D1 & \cmark &  62.77$^{\scalebox{0.6}{±1.60}}$ &           65.73$^{\scalebox{0.6}{±3.34}}$ &  \textbf{67.83$^{\scalebox{0.6}{±1.84}}$} &           67.56$^{\scalebox{0.6}{±1.99}}$ &  \textbf{68.47$^{\scalebox{0.6}{±1.53}}$} &  70.77$^{\scalebox{0.6}{±1.11}}$ \\
& D3$\rightarrow$D2 & \xmark &  56.29$^{\scalebox{0.6}{±1.58}}$ &           55.16$^{\scalebox{0.6}{±1.55}}$ &  \textbf{55.47$^{\scalebox{0.6}{±1.37}}$} &           55.84$^{\scalebox{0.6}{±1.62}}$ &  \textbf{56.03$^{\scalebox{0.6}{±1.56}}$} &  57.20$^{\scalebox{0.6}{±1.50}}$ \\
& D3$\rightarrow$D2 & \cmark &  56.29$^{\scalebox{0.6}{±1.59}}$ &           54.50$^{\scalebox{0.6}{±1.76}}$ &  \textbf{55.15$^{\scalebox{0.6}{±1.41}}$} &           55.67$^{\scalebox{0.6}{±1.36}}$ &  \textbf{56.06$^{\scalebox{0.6}{±1.42}}$} &  57.69$^{\scalebox{0.6}{±1.36}}$ \\
& D3$\rightarrow$D3 & \xmark &  54.30$^{\scalebox{0.6}{±1.16}}$ &           52.82$^{\scalebox{0.6}{±1.25}}$ &  \textbf{53.43$^{\scalebox{0.6}{±0.87}}$} &           53.31$^{\scalebox{0.6}{±1.03}}$ &  \textbf{53.59$^{\scalebox{0.6}{±1.03}}$} &  54.34$^{\scalebox{0.6}{±1.20}}$ \\
& D3$\rightarrow$D3 & \cmark &  54.31$^{\scalebox{0.6}{±1.16}}$ &           51.81$^{\scalebox{0.6}{±1.48}}$ &  \textbf{52.71$^{\scalebox{0.6}{±0.97}}$} &           52.96$^{\scalebox{0.6}{±1.01}}$ &  \textbf{53.41$^{\scalebox{0.6}{±1.08}}$} &  54.40$^{\scalebox{0.6}{±1.17}}$ \\
& D3$\rightarrow$D4 & \xmark &  53.55$^{\scalebox{0.6}{±1.03}}$ &           52.52$^{\scalebox{0.6}{±0.83}}$ &  \textbf{52.82$^{\scalebox{0.6}{±0.99}}$} &           52.87$^{\scalebox{0.6}{±0.95}}$ &  \textbf{53.12$^{\scalebox{0.6}{±0.94}}$} &  53.55$^{\scalebox{0.6}{±1.03}}$ \\
& D3$\rightarrow$D4 & \cmark &  53.55$^{\scalebox{0.6}{±1.03}}$ &           51.65$^{\scalebox{0.6}{±0.93}}$ &  \textbf{52.22$^{\scalebox{0.6}{±0.94}}$} &           52.41$^{\scalebox{0.6}{±0.93}}$ &  \textbf{52.96$^{\scalebox{0.6}{±0.95}}$} &  53.59$^{\scalebox{0.6}{±1.02}}$ \\
& D3$\rightarrow$D5 & \xmark &  49.10$^{\scalebox{0.6}{±0.81}}$ &           46.76$^{\scalebox{0.6}{±3.32}}$ &  \textbf{49.36$^{\scalebox{0.6}{±0.93}}$} &           49.22$^{\scalebox{0.6}{±1.36}}$ &  \textbf{50.15$^{\scalebox{0.6}{±0.89}}$} &  51.39$^{\scalebox{0.6}{±0.92}}$ \\
& D3$\rightarrow$D5 & \cmark &  49.10$^{\scalebox{0.6}{±0.82}}$ &           45.95$^{\scalebox{0.6}{±3.83}}$ &  \textbf{49.25$^{\scalebox{0.6}{±0.98}}$} &           49.52$^{\scalebox{0.6}{±1.58}}$ &  \textbf{50.75$^{\scalebox{0.6}{±0.99}}$} &  52.31$^{\scalebox{0.6}{±0.74}}$ \\
& D3$\rightarrow$UNI & \xmark &  51.10$^{\scalebox{0.6}{±0.97}}$ &           49.76$^{\scalebox{0.6}{±1.61}}$ &  \textbf{50.81$^{\scalebox{0.6}{±0.86}}$} &           50.70$^{\scalebox{0.6}{±1.02}}$ &  \textbf{51.08$^{\scalebox{0.6}{±0.98}}$} &  51.46$^{\scalebox{0.6}{±0.88}}$ \\
& D3$\rightarrow$UNI & \cmark &  51.10$^{\scalebox{0.6}{±0.97}}$ &           48.66$^{\scalebox{0.6}{±1.96}}$ &  \textbf{50.12$^{\scalebox{0.6}{±1.05}}$} &           50.33$^{\scalebox{0.6}{±0.96}}$ &  \textbf{51.04$^{\scalebox{0.6}{±1.00}}$} &  51.70$^{\scalebox{0.6}{±0.85}}$ \\
& UNI$\rightarrow$D1 & \xmark &  64.68$^{\scalebox{0.6}{±1.22}}$ &  \textbf{71.94$^{\scalebox{0.6}{±1.42}}$} &           71.86$^{\scalebox{0.6}{±1.63}}$ &           72.16$^{\scalebox{0.6}{±1.34}}$ &  \textbf{72.27$^{\scalebox{0.6}{±1.35}}$} &  74.62$^{\scalebox{0.6}{±1.11}}$ \\
& UNI$\rightarrow$D1 & \cmark &  64.68$^{\scalebox{0.6}{±1.22}}$ &           73.59$^{\scalebox{0.6}{±1.48}}$ &  \textbf{73.84$^{\scalebox{0.6}{±1.51}}$} &           74.36$^{\scalebox{0.6}{±1.22}}$ &  \textbf{74.45$^{\scalebox{0.6}{±1.29}}$} &  76.35$^{\scalebox{0.6}{±1.05}}$ \\
& UNI$\rightarrow$D2 & \xmark &  62.60$^{\scalebox{0.6}{±0.65}}$ &           64.47$^{\scalebox{0.6}{±0.71}}$ &  \textbf{64.53$^{\scalebox{0.6}{±0.70}}$} &           64.63$^{\scalebox{0.6}{±0.61}}$ &  \textbf{64.71$^{\scalebox{0.6}{±0.56}}$} &  65.28$^{\scalebox{0.6}{±0.57}}$ \\
& UNI$\rightarrow$D2 & \cmark &  62.61$^{\scalebox{0.6}{±0.66}}$ &           65.10$^{\scalebox{0.6}{±0.74}}$ &  \textbf{65.11$^{\scalebox{0.6}{±0.72}}$} &           65.26$^{\scalebox{0.6}{±0.63}}$ &  \textbf{65.33$^{\scalebox{0.6}{±0.61}}$} &  66.24$^{\scalebox{0.6}{±0.56}}$ \\
& UNI$\rightarrow$D3 & \xmark &  63.98$^{\scalebox{0.6}{±0.85}}$ &  \textbf{64.37$^{\scalebox{0.6}{±0.77}}$} &           64.37$^{\scalebox{0.6}{±0.77}}$ &           64.41$^{\scalebox{0.6}{±0.83}}$ &  \textbf{64.42$^{\scalebox{0.6}{±0.81}}$} &  64.75$^{\scalebox{0.6}{±0.81}}$ \\
& UNI$\rightarrow$D3 & \cmark &  63.98$^{\scalebox{0.6}{±0.85}}$ &           64.33$^{\scalebox{0.6}{±0.80}}$ &  \textbf{64.35$^{\scalebox{0.6}{±0.80}}$} &  \textbf{64.38$^{\scalebox{0.6}{±0.81}}$} &           64.38$^{\scalebox{0.6}{±0.81}}$ &  64.91$^{\scalebox{0.6}{±0.74}}$ \\
& UNI$\rightarrow$D4 & \xmark &  63.98$^{\scalebox{0.6}{±0.70}}$ &  \textbf{64.18$^{\scalebox{0.6}{±0.62}}$} &           64.17$^{\scalebox{0.6}{±0.61}}$ &  \textbf{64.24$^{\scalebox{0.6}{±0.71}}$} &           64.23$^{\scalebox{0.6}{±0.73}}$ &  64.35$^{\scalebox{0.6}{±0.63}}$ \\
& UNI$\rightarrow$D4 & \cmark &  63.98$^{\scalebox{0.6}{±0.70}}$ &  \textbf{64.20$^{\scalebox{0.6}{±0.66}}$} &           64.20$^{\scalebox{0.6}{±0.67}}$ &  \textbf{64.30$^{\scalebox{0.6}{±0.67}}$} &           64.29$^{\scalebox{0.6}{±0.66}}$ &  64.57$^{\scalebox{0.6}{±0.66}}$ \\
& UNI$\rightarrow$D5 & \xmark &  64.15$^{\scalebox{0.6}{±1.00}}$ &  \textbf{64.42$^{\scalebox{0.6}{±0.92}}$} &           64.42$^{\scalebox{0.6}{±0.93}}$ &  \textbf{64.58$^{\scalebox{0.6}{±0.94}}$} &           64.55$^{\scalebox{0.6}{±0.96}}$ &  64.91$^{\scalebox{0.6}{±0.92}}$ \\
& UNI$\rightarrow$D5 & \cmark &  64.15$^{\scalebox{0.6}{±0.99}}$ &           64.33$^{\scalebox{0.6}{±0.97}}$ &  \textbf{64.35$^{\scalebox{0.6}{±0.97}}$} &  \textbf{64.64$^{\scalebox{0.6}{±0.84}}$} &           64.63$^{\scalebox{0.6}{±0.82}}$ &  65.21$^{\scalebox{0.6}{±0.97}}$ \\
& UNI$\rightarrow$UNI & \xmark &  63.95$^{\scalebox{0.6}{±0.60}}$ &  \textbf{63.85$^{\scalebox{0.6}{±0.62}}$} &           63.85$^{\scalebox{0.6}{±0.62}}$ &           63.78$^{\scalebox{0.6}{±0.64}}$ &  \textbf{63.79$^{\scalebox{0.6}{±0.64}}$} &  63.96$^{\scalebox{0.6}{±0.60}}$ \\
& UNI$\rightarrow$UNI & \cmark &  63.96$^{\scalebox{0.6}{±0.59}}$ &  \textbf{63.69$^{\scalebox{0.6}{±0.68}}$} &           63.69$^{\scalebox{0.6}{±0.67}}$ &  \textbf{63.68$^{\scalebox{0.6}{±0.66}}$} &           63.68$^{\scalebox{0.6}{±0.67}}$ &  63.98$^{\scalebox{0.6}{±0.60}}$ \\

\hline
\multirow{24}{*}{ \rotatebox[origin=c]{90}{Places365}} 
& D3$\rightarrow$D1 & \xmark &  60.04$^{\scalebox{0.6}{±0.20}}$ &           64.68$^{\scalebox{0.6}{±0.94}}$ &  \textbf{65.71$^{\scalebox{0.6}{±0.11}}$} &           65.86$^{\scalebox{0.6}{±0.12}}$ &  \textbf{66.01$^{\scalebox{0.6}{±0.16}}$} &  66.75$^{\scalebox{0.6}{±0.14}}$ \\
& D3$\rightarrow$D1 & \cmark &  60.04$^{\scalebox{0.6}{±0.20}}$ &           64.69$^{\scalebox{0.6}{±0.97}}$ &  \textbf{65.75$^{\scalebox{0.6}{±0.13}}$} &           65.90$^{\scalebox{0.6}{±0.15}}$ &  \textbf{66.03$^{\scalebox{0.6}{±0.16}}$} &  66.73$^{\scalebox{0.6}{±0.14}}$ \\
& D3$\rightarrow$D2 & \xmark &  54.11$^{\scalebox{0.6}{±0.08}}$ &           53.20$^{\scalebox{0.6}{±0.86}}$ &  \textbf{54.26$^{\scalebox{0.6}{±0.09}}$} &           54.45$^{\scalebox{0.6}{±0.10}}$ &  \textbf{54.80$^{\scalebox{0.6}{±0.07}}$} &  55.16$^{\scalebox{0.6}{±0.05}}$ \\
& D3$\rightarrow$D2 & \cmark &  54.11$^{\scalebox{0.6}{±0.08}}$ &           53.17$^{\scalebox{0.6}{±0.89}}$ &  \textbf{54.25$^{\scalebox{0.6}{±0.09}}$} &           54.44$^{\scalebox{0.6}{±0.10}}$ &  \textbf{54.79$^{\scalebox{0.6}{±0.06}}$} &  55.17$^{\scalebox{0.6}{±0.05}}$ \\
& D3$\rightarrow$D3 & \xmark &  50.84$^{\scalebox{0.6}{±0.06}}$ &           48.56$^{\scalebox{0.6}{±1.17}}$ &  \textbf{50.13$^{\scalebox{0.6}{±0.09}}$} &           50.26$^{\scalebox{0.6}{±0.09}}$ &  \textbf{50.87$^{\scalebox{0.6}{±0.08}}$} &  50.99$^{\scalebox{0.6}{±0.05}}$ \\
& D3$\rightarrow$D3 & \cmark &  50.84$^{\scalebox{0.6}{±0.06}}$ &           48.50$^{\scalebox{0.6}{±1.19}}$ &  \textbf{50.09$^{\scalebox{0.6}{±0.09}}$} &           50.25$^{\scalebox{0.6}{±0.10}}$ &  \textbf{50.76$^{\scalebox{0.6}{±0.09}}$} &  50.99$^{\scalebox{0.6}{±0.05}}$ \\
& D3$\rightarrow$D4 & \xmark &  49.66$^{\scalebox{0.6}{±0.05}}$ &           48.24$^{\scalebox{0.6}{±0.38}}$ &  \textbf{49.19$^{\scalebox{0.6}{±0.10}}$} &           49.38$^{\scalebox{0.6}{±0.04}}$ &  \textbf{49.76$^{\scalebox{0.6}{±0.05}}$} &  49.97$^{\scalebox{0.6}{±0.08}}$ \\
& D3$\rightarrow$D4 & \cmark &  49.66$^{\scalebox{0.6}{±0.05}}$ &           48.19$^{\scalebox{0.6}{±0.39}}$ &  \textbf{49.18$^{\scalebox{0.6}{±0.09}}$} &           49.37$^{\scalebox{0.6}{±0.05}}$ &  \textbf{49.77$^{\scalebox{0.6}{±0.05}}$} &  49.98$^{\scalebox{0.6}{±0.07}}$ \\
& D3$\rightarrow$D5 & \xmark &  45.20$^{\scalebox{0.6}{±0.08}}$ &           47.06$^{\scalebox{0.6}{±1.31}}$ &  \textbf{48.98$^{\scalebox{0.6}{±0.20}}$} &           49.42$^{\scalebox{0.6}{±0.03}}$ &  \textbf{49.68$^{\scalebox{0.6}{±0.14}}$} &  50.28$^{\scalebox{0.6}{±0.14}}$ \\
& D3$\rightarrow$D5 & \cmark &  45.19$^{\scalebox{0.6}{±0.07}}$ &           46.99$^{\scalebox{0.6}{±1.36}}$ &  \textbf{48.95$^{\scalebox{0.6}{±0.21}}$} &           49.41$^{\scalebox{0.6}{±0.04}}$ &  \textbf{49.68$^{\scalebox{0.6}{±0.15}}$} &  50.31$^{\scalebox{0.6}{±0.14}}$ \\
& D3$\rightarrow$UNI & \xmark &  47.10$^{\scalebox{0.6}{±0.03}}$ &           46.36$^{\scalebox{0.6}{±1.11}}$ &  \textbf{47.74$^{\scalebox{0.6}{±0.08}}$} &  \textbf{48.08$^{\scalebox{0.6}{±0.06}}$} &           47.95$^{\scalebox{0.6}{±0.06}}$ &  48.64$^{\scalebox{0.6}{±0.03}}$ \\
& D3$\rightarrow$UNI & \cmark &  47.10$^{\scalebox{0.6}{±0.03}}$ &           46.29$^{\scalebox{0.6}{±1.14}}$ &  \textbf{47.72$^{\scalebox{0.6}{±0.09}}$} &  \textbf{48.08$^{\scalebox{0.6}{±0.06}}$} &           47.94$^{\scalebox{0.6}{±0.06}}$ &  48.65$^{\scalebox{0.6}{±0.03}}$ \\
& UNI$\rightarrow$D1 & \xmark &  53.12$^{\scalebox{0.6}{±0.09}}$ &           66.75$^{\scalebox{0.6}{±0.24}}$ &  \textbf{66.99$^{\scalebox{0.6}{±0.14}}$} &           67.25$^{\scalebox{0.6}{±0.16}}$ &  \textbf{67.29$^{\scalebox{0.6}{±0.14}}$} &  68.10$^{\scalebox{0.6}{±0.11}}$ \\
& UNI$\rightarrow$D1 & \cmark &  53.12$^{\scalebox{0.6}{±0.09}}$ &           66.75$^{\scalebox{0.6}{±0.24}}$ &  \textbf{66.98$^{\scalebox{0.6}{±0.15}}$} &           67.25$^{\scalebox{0.6}{±0.16}}$ &  \textbf{67.29$^{\scalebox{0.6}{±0.14}}$} &  68.10$^{\scalebox{0.6}{±0.11}}$ \\
& UNI$\rightarrow$D2 & \xmark &  52.29$^{\scalebox{0.6}{±0.06}}$ &           56.51$^{\scalebox{0.6}{±0.11}}$ &  \textbf{56.63$^{\scalebox{0.6}{±0.10}}$} &           56.88$^{\scalebox{0.6}{±0.09}}$ &  \textbf{56.95$^{\scalebox{0.6}{±0.08}}$} &  57.38$^{\scalebox{0.6}{±0.10}}$ \\
& UNI$\rightarrow$D2 & \cmark &  52.29$^{\scalebox{0.6}{±0.06}}$ &           56.50$^{\scalebox{0.6}{±0.11}}$ &  \textbf{56.63$^{\scalebox{0.6}{±0.10}}$} &           56.89$^{\scalebox{0.6}{±0.09}}$ &  \textbf{56.95$^{\scalebox{0.6}{±0.07}}$} &  57.37$^{\scalebox{0.6}{±0.11}}$ \\
& UNI$\rightarrow$D3 & \xmark &  52.10$^{\scalebox{0.6}{±0.07}}$ &           53.01$^{\scalebox{0.6}{±0.07}}$ &  \textbf{53.14$^{\scalebox{0.6}{±0.09}}$} &           53.37$^{\scalebox{0.6}{±0.10}}$ &  \textbf{53.50$^{\scalebox{0.6}{±0.07}}$} &  53.72$^{\scalebox{0.6}{±0.08}}$ \\
& UNI$\rightarrow$D3 & \cmark &  52.10$^{\scalebox{0.6}{±0.07}}$ &           53.01$^{\scalebox{0.6}{±0.07}}$ &  \textbf{53.14$^{\scalebox{0.6}{±0.09}}$} &           53.37$^{\scalebox{0.6}{±0.10}}$ &  \textbf{53.50$^{\scalebox{0.6}{±0.07}}$} &  53.72$^{\scalebox{0.6}{±0.08}}$ \\
& UNI$\rightarrow$D4 & \xmark &  51.68$^{\scalebox{0.6}{±0.10}}$ &           52.28$^{\scalebox{0.6}{±0.13}}$ &  \textbf{52.37$^{\scalebox{0.6}{±0.08}}$} &           52.51$^{\scalebox{0.6}{±0.08}}$ &  \textbf{52.61$^{\scalebox{0.6}{±0.07}}$} &  52.81$^{\scalebox{0.6}{±0.06}}$ \\
& UNI$\rightarrow$D4 & \cmark &  51.68$^{\scalebox{0.6}{±0.10}}$ &           52.28$^{\scalebox{0.6}{±0.13}}$ &  \textbf{52.37$^{\scalebox{0.6}{±0.08}}$} &           52.51$^{\scalebox{0.6}{±0.08}}$ &  \textbf{52.61$^{\scalebox{0.6}{±0.08}}$} &  52.81$^{\scalebox{0.6}{±0.05}}$ \\
& UNI$\rightarrow$D5 & \xmark &  52.49$^{\scalebox{0.6}{±0.15}}$ &           53.23$^{\scalebox{0.6}{±0.15}}$ &  \textbf{53.40$^{\scalebox{0.6}{±0.16}}$} &           53.59$^{\scalebox{0.6}{±0.10}}$ &  \textbf{53.78$^{\scalebox{0.6}{±0.06}}$} &  54.11$^{\scalebox{0.6}{±0.10}}$ \\
& UNI$\rightarrow$D5 & \cmark &  52.49$^{\scalebox{0.6}{±0.14}}$ &           53.23$^{\scalebox{0.6}{±0.15}}$ &  \textbf{53.40$^{\scalebox{0.6}{±0.16}}$} &           53.59$^{\scalebox{0.6}{±0.09}}$ &  \textbf{53.78$^{\scalebox{0.6}{±0.06}}$} &  54.11$^{\scalebox{0.6}{±0.10}}$ \\
& UNI$\rightarrow$UNI & \xmark &  51.77$^{\scalebox{0.6}{±0.09}}$ &           51.56$^{\scalebox{0.6}{±0.07}}$ &  \textbf{51.65$^{\scalebox{0.6}{±0.09}}$} &           51.75$^{\scalebox{0.6}{±0.07}}$ &  \textbf{51.89$^{\scalebox{0.6}{±0.07}}$} &  52.01$^{\scalebox{0.6}{±0.09}}$ \\
& UNI$\rightarrow$UNI & \cmark &  51.77$^{\scalebox{0.6}{±0.09}}$ &           51.56$^{\scalebox{0.6}{±0.07}}$ &  \textbf{51.65$^{\scalebox{0.6}{±0.09}}$} &           51.75$^{\scalebox{0.6}{±0.06}}$ &  \textbf{51.89$^{\scalebox{0.6}{±0.08}}$} &  52.01$^{\scalebox{0.6}{±0.08}}$ \\

\hline
\end{tabular}
\vspace*{.5mm}
\caption{\textit{``Improve Estimates from Confusion Matrix.'}' Accuracy (± std. dev.) after adaptation with new prior estimate based on confusion matrix (CM) inversion \cite{saerens2002adjusting} and our proposed method from Section \ref{subsubsection:cm_maximum_likelihood} (CM$^\text{L}$). SCM denotes soft confusion matrix, NA denotes no adaptation, Oracle is adaptation with ground truth priors. Results on CIFAR are averaged from 10 experiments, results on Places and ImageNet are averaged from 5 experiments. Best results are displayed in bold.}
\label{tab:cm_correction_sup} 
\end{table*}

\begin{table*}[hbt] 
\centering
\setlength\tabcolsep{3pt}
\hspace*{-1.55mm}\begin{tabular}{>{\centering\arraybackslash}m{1.cm}|m{1.6cm}||>{\centering\arraybackslash}m{1.35cm}||>{\centering\arraybackslash}m{1.375cm}|>{\centering\arraybackslash}m{1.375cm}>{\centering\arraybackslash}m{1.375cm}>{\centering\arraybackslash}m{1.375cm}|>{\centering\arraybackslash}m{1.375cm}>{\centering\arraybackslash}m{1.375cm} >{\centering\arraybackslash}m{1.375cm}|>{\centering\arraybackslash}m{1.375cm}}
\multicolumn{2}{c||}{} & \multirow{2}{*}{\shortstack{BCTS \cite{alexandari2020maximum}\\ calibrated}} & \multirow{2}{*}{NA} & \multicolumn{3}{c|}{MLE} & \multicolumn{3}{c|}{MAP}  & \multirow{2}{*}{Oracle} \\
\multicolumn{2}{c||}{Dataset} &  &   & EM & CM$^\text{L}$ & SCM$^\text{L}$ & MAP & CM$^\text{M}$ & SCM$^\text{M}$ \\
\hline \hline
 \multirow{24}{*}{ \rotatebox[origin=c]{90}{CIFAR-100}} 
& D3$\rightarrow$D1 & \xmark &  62.76$^{\scalebox{0.6}{±1.60}}$ &           65.63$^{\scalebox{0.6}{±1.58}}$ &           66.94$^{\scalebox{0.6}{±1.66}}$ &  \textbf{67.35$^{\scalebox{0.6}{±1.56}}$} &           64.64$^{\scalebox{0.6}{±1.78}}$ &           64.55$^{\scalebox{0.6}{±1.69}}$ &  \textbf{64.65$^{\scalebox{0.6}{±1.70}}$} &  69.18$^{\scalebox{0.6}{±1.36}}$ \\
& D3$\rightarrow$D1 & \cmark &  62.77$^{\scalebox{0.6}{±1.60}}$ &           67.36$^{\scalebox{0.6}{±1.68}}$ &           67.83$^{\scalebox{0.6}{±1.84}}$ &  \textbf{68.47$^{\scalebox{0.6}{±1.53}}$} &           65.47$^{\scalebox{0.6}{±1.73}}$ &  \textbf{65.57$^{\scalebox{0.6}{±1.87}}$} &           65.18$^{\scalebox{0.6}{±1.61}}$ &  70.77$^{\scalebox{0.6}{±1.11}}$ \\
& D3$\rightarrow$D2 & \xmark &  56.29$^{\scalebox{0.6}{±1.58}}$ &  \textbf{56.55$^{\scalebox{0.6}{±1.60}}$} &           55.47$^{\scalebox{0.6}{±1.37}}$ &           56.03$^{\scalebox{0.6}{±1.56}}$ &  \textbf{56.45$^{\scalebox{0.6}{±1.63}}$} &           56.30$^{\scalebox{0.6}{±1.54}}$ &           56.34$^{\scalebox{0.6}{±1.48}}$ &  57.20$^{\scalebox{0.6}{±1.50}}$ \\
& D3$\rightarrow$D2 & \cmark &  56.29$^{\scalebox{0.6}{±1.59}}$ &  \textbf{56.14$^{\scalebox{0.6}{±1.55}}$} &           55.15$^{\scalebox{0.6}{±1.41}}$ &           56.06$^{\scalebox{0.6}{±1.42}}$ &           56.10$^{\scalebox{0.6}{±1.56}}$ &  \textbf{56.22$^{\scalebox{0.6}{±1.52}}$} &           56.19$^{\scalebox{0.6}{±1.56}}$ &  57.69$^{\scalebox{0.6}{±1.36}}$ \\
& D3$\rightarrow$D3 & \xmark &  54.30$^{\scalebox{0.6}{±1.16}}$ &  \textbf{54.05$^{\scalebox{0.6}{±1.25}}$} &           53.43$^{\scalebox{0.6}{±0.87}}$ &           53.59$^{\scalebox{0.6}{±1.03}}$ &  \textbf{54.08$^{\scalebox{0.6}{±1.20}}$} &           53.95$^{\scalebox{0.6}{±1.11}}$ &           54.03$^{\scalebox{0.6}{±1.09}}$ &  54.34$^{\scalebox{0.6}{±1.20}}$ \\
& D3$\rightarrow$D3 & \cmark &  54.31$^{\scalebox{0.6}{±1.16}}$ &  \textbf{53.43$^{\scalebox{0.6}{±1.27}}$} &           52.71$^{\scalebox{0.6}{±0.97}}$ &           53.41$^{\scalebox{0.6}{±1.08}}$ &           53.67$^{\scalebox{0.6}{±1.25}}$ &           53.54$^{\scalebox{0.6}{±1.00}}$ &  \textbf{53.87$^{\scalebox{0.6}{±1.12}}$} &  54.40$^{\scalebox{0.6}{±1.17}}$ \\
& D3$\rightarrow$D4 & \xmark &  53.55$^{\scalebox{0.6}{±1.03}}$ &  \textbf{53.43$^{\scalebox{0.6}{±1.03}}$} &           52.82$^{\scalebox{0.6}{±0.99}}$ &           53.12$^{\scalebox{0.6}{±0.94}}$ &  \textbf{53.43$^{\scalebox{0.6}{±1.05}}$} &           53.30$^{\scalebox{0.6}{±1.02}}$ &           53.36$^{\scalebox{0.6}{±0.98}}$ &  53.55$^{\scalebox{0.6}{±1.03}}$ \\
& D3$\rightarrow$D4 & \cmark &  53.55$^{\scalebox{0.6}{±1.03}}$ &           52.91$^{\scalebox{0.6}{±0.98}}$ &           52.22$^{\scalebox{0.6}{±0.94}}$ &  \textbf{52.96$^{\scalebox{0.6}{±0.95}}$} &           53.01$^{\scalebox{0.6}{±1.00}}$ &           52.93$^{\scalebox{0.6}{±1.06}}$ &  \textbf{53.23$^{\scalebox{0.6}{±0.99}}$} &  53.59$^{\scalebox{0.6}{±1.02}}$ \\
& D3$\rightarrow$D5 & \xmark &  49.10$^{\scalebox{0.6}{±0.81}}$ &  \textbf{50.37$^{\scalebox{0.6}{±0.90}}$} &           49.36$^{\scalebox{0.6}{±0.93}}$ &           50.15$^{\scalebox{0.6}{±0.89}}$ &           50.33$^{\scalebox{0.6}{±0.94}}$ &  \textbf{50.86$^{\scalebox{0.6}{±0.86}}$} &           50.85$^{\scalebox{0.6}{±0.91}}$ &  51.39$^{\scalebox{0.6}{±0.92}}$ \\
& D3$\rightarrow$D5 & \cmark &  49.10$^{\scalebox{0.6}{±0.82}}$ &  \textbf{51.28$^{\scalebox{0.6}{±0.84}}$} &           49.25$^{\scalebox{0.6}{±0.98}}$ &           50.75$^{\scalebox{0.6}{±0.99}}$ &           51.34$^{\scalebox{0.6}{±0.85}}$ &           51.50$^{\scalebox{0.6}{±0.82}}$ &  \textbf{51.59$^{\scalebox{0.6}{±0.84}}$} &  52.31$^{\scalebox{0.6}{±0.74}}$ \\
& D3$\rightarrow$UNI & \xmark &  51.10$^{\scalebox{0.6}{±0.97}}$ &  \textbf{51.27$^{\scalebox{0.6}{±0.95}}$} &           50.81$^{\scalebox{0.6}{±0.86}}$ &           51.08$^{\scalebox{0.6}{±0.98}}$ &           51.28$^{\scalebox{0.6}{±0.94}}$ &           51.34$^{\scalebox{0.6}{±0.93}}$ &  \textbf{51.36$^{\scalebox{0.6}{±0.94}}$} &  51.46$^{\scalebox{0.6}{±0.88}}$ \\
& D3$\rightarrow$UNI & \cmark &  51.10$^{\scalebox{0.6}{±0.97}}$ &  \textbf{51.06$^{\scalebox{0.6}{±0.95}}$} &           50.12$^{\scalebox{0.6}{±1.05}}$ &           51.04$^{\scalebox{0.6}{±1.00}}$ &           51.11$^{\scalebox{0.6}{±0.95}}$ &           50.94$^{\scalebox{0.6}{±0.99}}$ &  \textbf{51.32$^{\scalebox{0.6}{±0.95}}$} &  51.70$^{\scalebox{0.6}{±0.85}}$ \\
& UNI$\rightarrow$D1 & \xmark &  64.68$^{\scalebox{0.6}{±1.22}}$ &           70.39$^{\scalebox{0.6}{±1.34}}$ &           71.86$^{\scalebox{0.6}{±1.63}}$ &  \textbf{72.27$^{\scalebox{0.6}{±1.35}}$} &           69.09$^{\scalebox{0.6}{±1.23}}$ &           69.44$^{\scalebox{0.6}{±1.38}}$ &  \textbf{69.48$^{\scalebox{0.6}{±1.29}}$} &  74.62$^{\scalebox{0.6}{±1.11}}$ \\
& UNI$\rightarrow$D1 & \cmark &  64.68$^{\scalebox{0.6}{±1.22}}$ &           73.40$^{\scalebox{0.6}{±1.37}}$ &           73.84$^{\scalebox{0.6}{±1.51}}$ &  \textbf{74.45$^{\scalebox{0.6}{±1.29}}$} &           71.48$^{\scalebox{0.6}{±1.59}}$ &  \textbf{71.58$^{\scalebox{0.6}{±1.39}}$} &           71.34$^{\scalebox{0.6}{±1.37}}$ &  76.35$^{\scalebox{0.6}{±1.05}}$ \\
& UNI$\rightarrow$D2 & \xmark &  62.60$^{\scalebox{0.6}{±0.65}}$ &           64.44$^{\scalebox{0.6}{±0.58}}$ &           64.53$^{\scalebox{0.6}{±0.70}}$ &  \textbf{64.71$^{\scalebox{0.6}{±0.56}}$} &           64.17$^{\scalebox{0.6}{±0.63}}$ &           64.51$^{\scalebox{0.6}{±0.60}}$ &  \textbf{64.54$^{\scalebox{0.6}{±0.62}}$} &  65.28$^{\scalebox{0.6}{±0.57}}$ \\
& UNI$\rightarrow$D2 & \cmark &  62.61$^{\scalebox{0.6}{±0.66}}$ &           65.13$^{\scalebox{0.6}{±0.67}}$ &           65.11$^{\scalebox{0.6}{±0.72}}$ &  \textbf{65.33$^{\scalebox{0.6}{±0.61}}$} &           64.90$^{\scalebox{0.6}{±0.55}}$ &           65.22$^{\scalebox{0.6}{±0.63}}$ &  \textbf{65.23$^{\scalebox{0.6}{±0.57}}$} &  66.24$^{\scalebox{0.6}{±0.56}}$ \\
& UNI$\rightarrow$D3 & \xmark &  63.98$^{\scalebox{0.6}{±0.85}}$ &           64.38$^{\scalebox{0.6}{±0.90}}$ &           64.37$^{\scalebox{0.6}{±0.77}}$ &  \textbf{64.42$^{\scalebox{0.6}{±0.81}}$} &           64.33$^{\scalebox{0.6}{±0.92}}$ &           64.46$^{\scalebox{0.6}{±0.84}}$ &  \textbf{64.47$^{\scalebox{0.6}{±0.84}}$} &  64.75$^{\scalebox{0.6}{±0.81}}$ \\
& UNI$\rightarrow$D3 & \cmark &  63.98$^{\scalebox{0.6}{±0.85}}$ &  \textbf{64.43$^{\scalebox{0.6}{±0.83}}$} &           64.35$^{\scalebox{0.6}{±0.80}}$ &           64.38$^{\scalebox{0.6}{±0.81}}$ &           64.43$^{\scalebox{0.6}{±0.84}}$ &  \textbf{64.54$^{\scalebox{0.6}{±0.82}}$} &           64.54$^{\scalebox{0.6}{±0.83}}$ &  64.91$^{\scalebox{0.6}{±0.74}}$ \\
& UNI$\rightarrow$D4 & \xmark &  63.98$^{\scalebox{0.6}{±0.70}}$ &           64.14$^{\scalebox{0.6}{±0.72}}$ &           64.17$^{\scalebox{0.6}{±0.61}}$ &  \textbf{64.23$^{\scalebox{0.6}{±0.73}}$} &           64.15$^{\scalebox{0.6}{±0.72}}$ &           64.20$^{\scalebox{0.6}{±0.63}}$ &  \textbf{64.22$^{\scalebox{0.6}{±0.71}}$} &  64.35$^{\scalebox{0.6}{±0.63}}$ \\
& UNI$\rightarrow$D4 & \cmark &  63.98$^{\scalebox{0.6}{±0.70}}$ &           64.27$^{\scalebox{0.6}{±0.61}}$ &           64.20$^{\scalebox{0.6}{±0.67}}$ &  \textbf{64.29$^{\scalebox{0.6}{±0.66}}$} &           64.25$^{\scalebox{0.6}{±0.67}}$ &           64.31$^{\scalebox{0.6}{±0.65}}$ &  \textbf{64.36$^{\scalebox{0.6}{±0.68}}$} &  64.57$^{\scalebox{0.6}{±0.66}}$ \\
& UNI$\rightarrow$D5 & \xmark &  64.15$^{\scalebox{0.6}{±1.00}}$ &  \textbf{64.59$^{\scalebox{0.6}{±0.97}}$} &           64.42$^{\scalebox{0.6}{±0.93}}$ &           64.55$^{\scalebox{0.6}{±0.96}}$ &           64.52$^{\scalebox{0.6}{±1.01}}$ &  \textbf{64.66$^{\scalebox{0.6}{±0.88}}$} &           64.65$^{\scalebox{0.6}{±0.92}}$ &  64.91$^{\scalebox{0.6}{±0.92}}$ \\
& UNI$\rightarrow$D5 & \cmark &  64.15$^{\scalebox{0.6}{±0.99}}$ &  \textbf{64.71$^{\scalebox{0.6}{±0.94}}$} &           64.35$^{\scalebox{0.6}{±0.97}}$ &           64.63$^{\scalebox{0.6}{±0.82}}$ &           64.62$^{\scalebox{0.6}{±0.92}}$ &           64.76$^{\scalebox{0.6}{±0.93}}$ &  \textbf{64.82$^{\scalebox{0.6}{±0.95}}$} &  65.21$^{\scalebox{0.6}{±0.97}}$ \\
& UNI$\rightarrow$UNI & \xmark &  63.95$^{\scalebox{0.6}{±0.60}}$ &           63.81$^{\scalebox{0.6}{±0.62}}$ &  \textbf{63.85$^{\scalebox{0.6}{±0.62}}$} &           63.79$^{\scalebox{0.6}{±0.64}}$ &           63.82$^{\scalebox{0.6}{±0.62}}$ &  \textbf{63.86$^{\scalebox{0.6}{±0.59}}$} &           63.83$^{\scalebox{0.6}{±0.66}}$ &  63.96$^{\scalebox{0.6}{±0.60}}$ \\
& UNI$\rightarrow$UNI & \cmark &  63.96$^{\scalebox{0.6}{±0.59}}$ &           63.68$^{\scalebox{0.6}{±0.64}}$ &  \textbf{63.69$^{\scalebox{0.6}{±0.67}}$} &           63.68$^{\scalebox{0.6}{±0.67}}$ &           63.68$^{\scalebox{0.6}{±0.64}}$ &  \textbf{63.75$^{\scalebox{0.6}{±0.71}}$} &           63.74$^{\scalebox{0.6}{±0.66}}$ &  63.98$^{\scalebox{0.6}{±0.60}}$ \\

\hline
\multirow{24}{*}{ \rotatebox[origin=c]{90}{Places365}} 
& D3$\rightarrow$D1 & \xmark &  60.04$^{\scalebox{0.6}{±0.20}}$ &           65.83$^{\scalebox{0.6}{±0.18}}$ &           65.71$^{\scalebox{0.6}{±0.11}}$ &  \textbf{66.01$^{\scalebox{0.6}{±0.16}}$} &           62.68$^{\scalebox{0.6}{±0.19}}$ &           63.21$^{\scalebox{0.6}{±0.16}}$ &  \textbf{64.38$^{\scalebox{0.6}{±0.10}}$} &  66.75$^{\scalebox{0.6}{±0.14}}$ \\
& D3$\rightarrow$D1 & \cmark &  60.04$^{\scalebox{0.6}{±0.20}}$ &           65.86$^{\scalebox{0.6}{±0.17}}$ &           65.75$^{\scalebox{0.6}{±0.13}}$ &  \textbf{66.03$^{\scalebox{0.6}{±0.16}}$} &           62.71$^{\scalebox{0.6}{±0.19}}$ &           63.29$^{\scalebox{0.6}{±0.15}}$ &  \textbf{64.38$^{\scalebox{0.6}{±0.08}}$} &  66.73$^{\scalebox{0.6}{±0.14}}$ \\
& D3$\rightarrow$D2 & \xmark &  54.11$^{\scalebox{0.6}{±0.08}}$ &  \textbf{54.83$^{\scalebox{0.6}{±0.06}}$} &           54.26$^{\scalebox{0.6}{±0.09}}$ &           54.80$^{\scalebox{0.6}{±0.07}}$ &  \textbf{54.75$^{\scalebox{0.6}{±0.06}}$} &           54.71$^{\scalebox{0.6}{±0.07}}$ &           54.75$^{\scalebox{0.6}{±0.10}}$ &  55.16$^{\scalebox{0.6}{±0.05}}$ \\
& D3$\rightarrow$D2 & \cmark &  54.11$^{\scalebox{0.6}{±0.08}}$ &  \textbf{54.84$^{\scalebox{0.6}{±0.07}}$} &           54.25$^{\scalebox{0.6}{±0.09}}$ &           54.79$^{\scalebox{0.6}{±0.06}}$ &           54.74$^{\scalebox{0.6}{±0.05}}$ &           54.71$^{\scalebox{0.6}{±0.07}}$ &  \textbf{54.76$^{\scalebox{0.6}{±0.09}}$} &  55.17$^{\scalebox{0.6}{±0.05}}$ \\
& D3$\rightarrow$D3 & \xmark &  50.84$^{\scalebox{0.6}{±0.06}}$ &           50.82$^{\scalebox{0.6}{±0.04}}$ &           50.13$^{\scalebox{0.6}{±0.09}}$ &  \textbf{50.87$^{\scalebox{0.6}{±0.08}}$} &           50.54$^{\scalebox{0.6}{±0.06}}$ &           50.55$^{\scalebox{0.6}{±0.08}}$ &  \textbf{50.90$^{\scalebox{0.6}{±0.04}}$} &  50.99$^{\scalebox{0.6}{±0.05}}$ \\
& D3$\rightarrow$D3 & \cmark &  50.84$^{\scalebox{0.6}{±0.06}}$ &  \textbf{50.82$^{\scalebox{0.6}{±0.04}}$} &           50.09$^{\scalebox{0.6}{±0.09}}$ &           50.76$^{\scalebox{0.6}{±0.09}}$ &           50.53$^{\scalebox{0.6}{±0.07}}$ &           50.54$^{\scalebox{0.6}{±0.08}}$ &  \textbf{50.91$^{\scalebox{0.6}{±0.04}}$} &  50.99$^{\scalebox{0.6}{±0.05}}$ \\
& D3$\rightarrow$D4 & \xmark &  49.66$^{\scalebox{0.6}{±0.05}}$ &           49.72$^{\scalebox{0.6}{±0.07}}$ &           49.19$^{\scalebox{0.6}{±0.10}}$ &  \textbf{49.76$^{\scalebox{0.6}{±0.05}}$} &           49.69$^{\scalebox{0.6}{±0.07}}$ &           49.60$^{\scalebox{0.6}{±0.08}}$ &  \textbf{49.77$^{\scalebox{0.6}{±0.06}}$} &  49.97$^{\scalebox{0.6}{±0.08}}$ \\
& D3$\rightarrow$D4 & \cmark &  49.66$^{\scalebox{0.6}{±0.05}}$ &           49.72$^{\scalebox{0.6}{±0.06}}$ &           49.18$^{\scalebox{0.6}{±0.09}}$ &  \textbf{49.77$^{\scalebox{0.6}{±0.05}}$} &           49.69$^{\scalebox{0.6}{±0.07}}$ &           49.59$^{\scalebox{0.6}{±0.09}}$ &  \textbf{49.78$^{\scalebox{0.6}{±0.05}}$} &  49.98$^{\scalebox{0.6}{±0.07}}$ \\
& D3$\rightarrow$D5 & \xmark &  45.20$^{\scalebox{0.6}{±0.08}}$ &           49.60$^{\scalebox{0.6}{±0.07}}$ &           48.98$^{\scalebox{0.6}{±0.20}}$ &  \textbf{49.68$^{\scalebox{0.6}{±0.14}}$} &  \textbf{49.74$^{\scalebox{0.6}{±0.07}}$} &           49.63$^{\scalebox{0.6}{±0.11}}$ &           49.65$^{\scalebox{0.6}{±0.11}}$ &  50.28$^{\scalebox{0.6}{±0.14}}$ \\
& D3$\rightarrow$D5 & \cmark &  45.19$^{\scalebox{0.6}{±0.07}}$ &           49.66$^{\scalebox{0.6}{±0.08}}$ &           48.95$^{\scalebox{0.6}{±0.21}}$ &  \textbf{49.68$^{\scalebox{0.6}{±0.15}}$} &  \textbf{49.76$^{\scalebox{0.6}{±0.08}}$} &           49.63$^{\scalebox{0.6}{±0.11}}$ &           49.64$^{\scalebox{0.6}{±0.11}}$ &  50.31$^{\scalebox{0.6}{±0.14}}$ \\
& D3$\rightarrow$UNI & \xmark &  47.10$^{\scalebox{0.6}{±0.03}}$ &  \textbf{48.24$^{\scalebox{0.6}{±0.04}}$} &           47.74$^{\scalebox{0.6}{±0.08}}$ &           47.95$^{\scalebox{0.6}{±0.06}}$ &  \textbf{48.24$^{\scalebox{0.6}{±0.05}}$} &           48.23$^{\scalebox{0.6}{±0.03}}$ &           48.02$^{\scalebox{0.6}{±0.06}}$ &  48.64$^{\scalebox{0.6}{±0.03}}$ \\
& D3$\rightarrow$UNI & \cmark &  47.10$^{\scalebox{0.6}{±0.03}}$ &  \textbf{48.25$^{\scalebox{0.6}{±0.04}}$} &           47.72$^{\scalebox{0.6}{±0.09}}$ &           47.94$^{\scalebox{0.6}{±0.06}}$ &  \textbf{48.24$^{\scalebox{0.6}{±0.05}}$} &           48.21$^{\scalebox{0.6}{±0.02}}$ &           48.00$^{\scalebox{0.6}{±0.06}}$ &  48.65$^{\scalebox{0.6}{±0.03}}$ \\
& UNI$\rightarrow$D1 & \xmark &  53.12$^{\scalebox{0.6}{±0.09}}$ &           67.00$^{\scalebox{0.6}{±0.15}}$ &           66.99$^{\scalebox{0.6}{±0.14}}$ &  \textbf{67.29$^{\scalebox{0.6}{±0.14}}$} &           62.50$^{\scalebox{0.6}{±0.17}}$ &           62.48$^{\scalebox{0.6}{±0.18}}$ &  \textbf{65.32$^{\scalebox{0.6}{±0.13}}$} &  68.10$^{\scalebox{0.6}{±0.11}}$ \\
& UNI$\rightarrow$D1 & \cmark &  53.12$^{\scalebox{0.6}{±0.09}}$ &           67.01$^{\scalebox{0.6}{±0.15}}$ &           66.98$^{\scalebox{0.6}{±0.15}}$ &  \textbf{67.29$^{\scalebox{0.6}{±0.14}}$} &           62.51$^{\scalebox{0.6}{±0.16}}$ &           63.31$^{\scalebox{0.6}{±1.75}}$ &  \textbf{65.33$^{\scalebox{0.6}{±0.13}}$} &  68.10$^{\scalebox{0.6}{±0.11}}$ \\
& UNI$\rightarrow$D2 & \xmark &  52.29$^{\scalebox{0.6}{±0.06}}$ &           56.79$^{\scalebox{0.6}{±0.11}}$ &           56.63$^{\scalebox{0.6}{±0.10}}$ &  \textbf{56.95$^{\scalebox{0.6}{±0.08}}$} &  \textbf{56.88$^{\scalebox{0.6}{±0.08}}$} &           56.87$^{\scalebox{0.6}{±0.10}}$ &           56.71$^{\scalebox{0.6}{±0.13}}$ &  57.38$^{\scalebox{0.6}{±0.10}}$ \\
& UNI$\rightarrow$D2 & \cmark &  52.29$^{\scalebox{0.6}{±0.06}}$ &           56.79$^{\scalebox{0.6}{±0.11}}$ &           56.63$^{\scalebox{0.6}{±0.10}}$ &  \textbf{56.95$^{\scalebox{0.6}{±0.07}}$} &           56.69$^{\scalebox{0.6}{±0.42}}$ &  \textbf{56.88$^{\scalebox{0.6}{±0.10}}$} &           56.71$^{\scalebox{0.6}{±0.14}}$ &  57.37$^{\scalebox{0.6}{±0.11}}$ \\
& UNI$\rightarrow$D3 & \xmark &  52.10$^{\scalebox{0.6}{±0.07}}$ &           53.36$^{\scalebox{0.6}{±0.11}}$ &           53.14$^{\scalebox{0.6}{±0.09}}$ &  \textbf{53.50$^{\scalebox{0.6}{±0.07}}$} &           53.33$^{\scalebox{0.6}{±0.09}}$ &           53.38$^{\scalebox{0.6}{±0.08}}$ &  \textbf{53.51$^{\scalebox{0.6}{±0.08}}$} &  53.72$^{\scalebox{0.6}{±0.08}}$ \\
& UNI$\rightarrow$D3 & \cmark &  52.10$^{\scalebox{0.6}{±0.07}}$ &           53.36$^{\scalebox{0.6}{±0.11}}$ &           53.14$^{\scalebox{0.6}{±0.09}}$ &  \textbf{53.50$^{\scalebox{0.6}{±0.07}}$} &           53.33$^{\scalebox{0.6}{±0.10}}$ &           53.38$^{\scalebox{0.6}{±0.08}}$ &  \textbf{53.51$^{\scalebox{0.6}{±0.08}}$} &  53.72$^{\scalebox{0.6}{±0.08}}$ \\
& UNI$\rightarrow$D4 & \xmark &  51.68$^{\scalebox{0.6}{±0.10}}$ &           52.50$^{\scalebox{0.6}{±0.08}}$ &           52.37$^{\scalebox{0.6}{±0.08}}$ &  \textbf{52.61$^{\scalebox{0.6}{±0.07}}$} &           52.42$^{\scalebox{0.6}{±0.10}}$ &           52.52$^{\scalebox{0.6}{±0.06}}$ &  \textbf{52.62$^{\scalebox{0.6}{±0.09}}$} &  52.81$^{\scalebox{0.6}{±0.06}}$ \\
& UNI$\rightarrow$D4 & \cmark &  51.68$^{\scalebox{0.6}{±0.10}}$ &           52.50$^{\scalebox{0.6}{±0.09}}$ &           52.37$^{\scalebox{0.6}{±0.08}}$ &  \textbf{52.61$^{\scalebox{0.6}{±0.08}}$} &           52.42$^{\scalebox{0.6}{±0.09}}$ &           52.52$^{\scalebox{0.6}{±0.06}}$ &  \textbf{52.62$^{\scalebox{0.6}{±0.09}}$} &  52.81$^{\scalebox{0.6}{±0.05}}$ \\
& UNI$\rightarrow$D5 & \xmark &  52.49$^{\scalebox{0.6}{±0.15}}$ &           53.71$^{\scalebox{0.6}{±0.12}}$ &           53.40$^{\scalebox{0.6}{±0.16}}$ &  \textbf{53.78$^{\scalebox{0.6}{±0.06}}$} &           53.70$^{\scalebox{0.6}{±0.10}}$ &           53.72$^{\scalebox{0.6}{±0.08}}$ &  \textbf{53.77$^{\scalebox{0.6}{±0.13}}$} &  54.11$^{\scalebox{0.6}{±0.10}}$ \\
& UNI$\rightarrow$D5 & \cmark &  52.49$^{\scalebox{0.6}{±0.14}}$ &           53.71$^{\scalebox{0.6}{±0.12}}$ &           53.40$^{\scalebox{0.6}{±0.16}}$ &  \textbf{53.78$^{\scalebox{0.6}{±0.06}}$} &           53.70$^{\scalebox{0.6}{±0.10}}$ &           53.72$^{\scalebox{0.6}{±0.08}}$ &  \textbf{53.77$^{\scalebox{0.6}{±0.13}}$} &  54.11$^{\scalebox{0.6}{±0.10}}$ \\
& UNI$\rightarrow$UNI & \xmark &  51.77$^{\scalebox{0.6}{±0.09}}$ &           51.78$^{\scalebox{0.6}{±0.08}}$ &           51.65$^{\scalebox{0.6}{±0.09}}$ &  \textbf{51.89$^{\scalebox{0.6}{±0.07}}$} &           51.80$^{\scalebox{0.6}{±0.09}}$ &           51.80$^{\scalebox{0.6}{±0.09}}$ &  \textbf{51.89$^{\scalebox{0.6}{±0.08}}$} &  52.01$^{\scalebox{0.6}{±0.09}}$ \\
& UNI$\rightarrow$UNI & \cmark &  51.77$^{\scalebox{0.6}{±0.09}}$ &           51.78$^{\scalebox{0.6}{±0.08}}$ &           51.65$^{\scalebox{0.6}{±0.09}}$ &  \textbf{51.89$^{\scalebox{0.6}{±0.08}}$} &           51.80$^{\scalebox{0.6}{±0.09}}$ &           51.81$^{\scalebox{0.6}{±0.09}}$ &  \textbf{51.89$^{\scalebox{0.6}{±0.08}}$} &  52.01$^{\scalebox{0.6}{±0.08}}$ \\

\hline
\end{tabular}
\vspace*{.5mm}
\caption{\textit{``How to estimate new priors?''} Accuracy (± std. dev.) after adaptation to new priors estimated with different Maximum Likelihood and Maximum A Posteriori estimates. NA denotes no adaptation, Oracle is adaptation with ground truth priors. Best MLE and MAP results are underlined for $f_\mathcal{T}$ and calibrated $f_\mathcal{T}$. Results on CIFAR are averaged from 10 experiments, results on Places and ImageNet are averaged from 5 experiments. Best results are displayed in bold.}
\label{tab:mle_map_cm_all_sup} 
\end{table*}

\begin{table*}[h] 
\centering
\setlength\tabcolsep{3pt}
\hspace*{-1.55mm}\begin{tabular}{>{\centering\arraybackslash}m{1.cm}|m{1.6cm}||>{\centering\arraybackslash}m{1.35cm}||>{\centering\arraybackslash}m{1.90cm}|>{\centering\arraybackslash}m{1.90cm}>{\centering\arraybackslash}m{1.90cm}|>{\centering\arraybackslash}m{1.90cm}>{\centering\arraybackslash}m{1.90cm}|>{\centering\arraybackslash}m{1.90cm}}
\multicolumn{2}{c||}{Dataset} & \shortstack{BCTS \cite{alexandari2020maximum}\\ calibrated} & NA & SCM$^\text{L}$ & RLLS & BBSE & BBSE-S & Oracle \\
\hline \hline
 \multirow{24}{*}{ \rotatebox[origin=c]{90}{CIFAR-100}} 
& D3$\rightarrow$D1 & \xmark &  62.76$^{\scalebox{0.6}{±1.60}}$ &  \textbf{67.35$^{\scalebox{0.6}{±1.56}}$} &           67.08$^{\scalebox{0.6}{±1.63}}$ &  64.75$^{\scalebox{0.6}{±2.31}}$ &           67.02$^{\scalebox{0.6}{±1.80}}$ &  69.18$^{\scalebox{0.6}{±1.36}}$ \\
& D3$\rightarrow$D1 & \cmark &  62.77$^{\scalebox{0.6}{±1.60}}$ &           68.47$^{\scalebox{0.6}{±1.53}}$ &           68.03$^{\scalebox{0.6}{±1.71}}$ &  65.10$^{\scalebox{0.6}{±3.18}}$ &  \textbf{68.96$^{\scalebox{0.6}{±1.46}}$} &  70.77$^{\scalebox{0.6}{±1.11}}$ \\
& D3$\rightarrow$D2 & \xmark &  56.29$^{\scalebox{0.6}{±1.58}}$ &           56.03$^{\scalebox{0.6}{±1.56}}$ &           55.59$^{\scalebox{0.6}{±1.36}}$ &  56.00$^{\scalebox{0.6}{±1.50}}$ &  \textbf{57.02$^{\scalebox{0.6}{±1.45}}$} &  57.20$^{\scalebox{0.6}{±1.50}}$ \\
& D3$\rightarrow$D2 & \cmark &  56.29$^{\scalebox{0.6}{±1.59}}$ &           56.06$^{\scalebox{0.6}{±1.42}}$ &           55.21$^{\scalebox{0.6}{±1.44}}$ &  55.15$^{\scalebox{0.6}{±1.80}}$ &  \textbf{56.69$^{\scalebox{0.6}{±1.34}}$} &  57.69$^{\scalebox{0.6}{±1.36}}$ \\
& D3$\rightarrow$D3 & \xmark &  54.30$^{\scalebox{0.6}{±1.16}}$ &           53.59$^{\scalebox{0.6}{±1.03}}$ &           53.39$^{\scalebox{0.6}{±0.96}}$ &  53.56$^{\scalebox{0.6}{±1.06}}$ &  \textbf{53.92$^{\scalebox{0.6}{±0.86}}$} &  54.34$^{\scalebox{0.6}{±1.20}}$ \\
& D3$\rightarrow$D3 & \cmark &  54.31$^{\scalebox{0.6}{±1.16}}$ &  \textbf{53.41$^{\scalebox{0.6}{±1.08}}$} &           52.61$^{\scalebox{0.6}{±1.12}}$ &  52.36$^{\scalebox{0.6}{±1.28}}$ &           53.36$^{\scalebox{0.6}{±0.91}}$ &  54.40$^{\scalebox{0.6}{±1.17}}$ \\
& D3$\rightarrow$D4 & \xmark &  53.55$^{\scalebox{0.6}{±1.03}}$ &           53.12$^{\scalebox{0.6}{±0.94}}$ &           52.76$^{\scalebox{0.6}{±0.91}}$ &  52.97$^{\scalebox{0.6}{±0.90}}$ &  \textbf{53.21$^{\scalebox{0.6}{±0.86}}$} &  53.55$^{\scalebox{0.6}{±1.03}}$ \\
& D3$\rightarrow$D4 & \cmark &  53.55$^{\scalebox{0.6}{±1.03}}$ &  \textbf{52.96$^{\scalebox{0.6}{±0.95}}$} &           52.10$^{\scalebox{0.6}{±0.97}}$ &  51.97$^{\scalebox{0.6}{±0.90}}$ &           52.70$^{\scalebox{0.6}{±0.80}}$ &  53.59$^{\scalebox{0.6}{±1.02}}$ \\
& D3$\rightarrow$D5 & \xmark &  49.10$^{\scalebox{0.6}{±0.81}}$ &  \textbf{50.15$^{\scalebox{0.6}{±0.89}}$} &           49.17$^{\scalebox{0.6}{±1.02}}$ &  49.98$^{\scalebox{0.6}{±1.24}}$ &           49.84$^{\scalebox{0.6}{±0.97}}$ &  51.39$^{\scalebox{0.6}{±0.92}}$ \\
& D3$\rightarrow$D5 & \cmark &  49.10$^{\scalebox{0.6}{±0.82}}$ &  \textbf{50.75$^{\scalebox{0.6}{±0.99}}$} &           49.17$^{\scalebox{0.6}{±1.02}}$ &  48.59$^{\scalebox{0.6}{±1.63}}$ &           49.38$^{\scalebox{0.6}{±1.16}}$ &  52.31$^{\scalebox{0.6}{±0.74}}$ \\
& D3$\rightarrow$UNI & \xmark &  51.10$^{\scalebox{0.6}{±0.97}}$ &  \textbf{51.08$^{\scalebox{0.6}{±0.98}}$} &           50.71$^{\scalebox{0.6}{±0.91}}$ &  50.82$^{\scalebox{0.6}{±0.90}}$ &           50.73$^{\scalebox{0.6}{±0.98}}$ &  51.46$^{\scalebox{0.6}{±0.88}}$ \\
& D3$\rightarrow$UNI & \cmark &  51.10$^{\scalebox{0.6}{±0.97}}$ &  \textbf{51.04$^{\scalebox{0.6}{±1.00}}$} &           50.03$^{\scalebox{0.6}{±1.11}}$ &  49.41$^{\scalebox{0.6}{±1.35}}$ &           50.20$^{\scalebox{0.6}{±0.96}}$ &  51.70$^{\scalebox{0.6}{±0.85}}$ \\
& UNI$\rightarrow$D1 & \xmark &  64.68$^{\scalebox{0.6}{±1.22}}$ &  \textbf{72.27$^{\scalebox{0.6}{±1.35}}$} &           71.94$^{\scalebox{0.6}{±1.63}}$ &  71.05$^{\scalebox{0.6}{±1.37}}$ &           71.41$^{\scalebox{0.6}{±1.36}}$ &  74.62$^{\scalebox{0.6}{±1.11}}$ \\
& UNI$\rightarrow$D1 & \cmark &  64.68$^{\scalebox{0.6}{±1.22}}$ &  \textbf{74.45$^{\scalebox{0.6}{±1.29}}$} &           73.75$^{\scalebox{0.6}{±1.66}}$ &  73.29$^{\scalebox{0.6}{±1.40}}$ &           74.06$^{\scalebox{0.6}{±1.13}}$ &  76.35$^{\scalebox{0.6}{±1.05}}$ \\
& UNI$\rightarrow$D2 & \xmark &  62.60$^{\scalebox{0.6}{±0.65}}$ &           64.71$^{\scalebox{0.6}{±0.56}}$ &           64.51$^{\scalebox{0.6}{±0.68}}$ &  64.77$^{\scalebox{0.6}{±0.62}}$ &  \textbf{64.82$^{\scalebox{0.6}{±0.60}}$} &  65.28$^{\scalebox{0.6}{±0.57}}$ \\
& UNI$\rightarrow$D2 & \cmark &  62.61$^{\scalebox{0.6}{±0.66}}$ &           65.33$^{\scalebox{0.6}{±0.61}}$ &           65.11$^{\scalebox{0.6}{±0.71}}$ &  65.35$^{\scalebox{0.6}{±0.61}}$ &  \textbf{65.40$^{\scalebox{0.6}{±0.61}}$} &  66.24$^{\scalebox{0.6}{±0.56}}$ \\
& UNI$\rightarrow$D3 & \xmark &  63.98$^{\scalebox{0.6}{±0.85}}$ &           64.42$^{\scalebox{0.6}{±0.81}}$ &           64.36$^{\scalebox{0.6}{±0.77}}$ &  64.41$^{\scalebox{0.6}{±0.78}}$ &  \textbf{64.43$^{\scalebox{0.6}{±0.83}}$} &  64.75$^{\scalebox{0.6}{±0.81}}$ \\
& UNI$\rightarrow$D3 & \cmark &  63.98$^{\scalebox{0.6}{±0.85}}$ &           64.38$^{\scalebox{0.6}{±0.81}}$ &           64.35$^{\scalebox{0.6}{±0.80}}$ &  64.36$^{\scalebox{0.6}{±0.81}}$ &  \textbf{64.39$^{\scalebox{0.6}{±0.81}}$} &  64.91$^{\scalebox{0.6}{±0.74}}$ \\
& UNI$\rightarrow$D4 & \xmark &  63.98$^{\scalebox{0.6}{±0.70}}$ &           64.23$^{\scalebox{0.6}{±0.73}}$ &           64.18$^{\scalebox{0.6}{±0.62}}$ &  64.19$^{\scalebox{0.6}{±0.62}}$ &  \textbf{64.25$^{\scalebox{0.6}{±0.70}}$} &  64.35$^{\scalebox{0.6}{±0.63}}$ \\
& UNI$\rightarrow$D4 & \cmark &  63.98$^{\scalebox{0.6}{±0.70}}$ &           64.29$^{\scalebox{0.6}{±0.66}}$ &           64.20$^{\scalebox{0.6}{±0.67}}$ &  64.22$^{\scalebox{0.6}{±0.66}}$ &  \textbf{64.30$^{\scalebox{0.6}{±0.67}}$} &  64.57$^{\scalebox{0.6}{±0.66}}$ \\
& UNI$\rightarrow$D5 & \xmark &  64.15$^{\scalebox{0.6}{±1.00}}$ &           64.55$^{\scalebox{0.6}{±0.96}}$ &           64.43$^{\scalebox{0.6}{±0.92}}$ &  64.58$^{\scalebox{0.6}{±0.95}}$ &  \textbf{64.67$^{\scalebox{0.6}{±0.97}}$} &  64.91$^{\scalebox{0.6}{±0.92}}$ \\
& UNI$\rightarrow$D5 & \cmark &  64.15$^{\scalebox{0.6}{±0.99}}$ &           64.63$^{\scalebox{0.6}{±0.82}}$ &           64.35$^{\scalebox{0.6}{±0.98}}$ &  64.44$^{\scalebox{0.6}{±0.98}}$ &  \textbf{64.68$^{\scalebox{0.6}{±0.87}}$} &  65.21$^{\scalebox{0.6}{±0.97}}$ \\
& UNI$\rightarrow$UNI & \xmark &  63.95$^{\scalebox{0.6}{±0.60}}$ &           63.79$^{\scalebox{0.6}{±0.64}}$ &  \textbf{63.85$^{\scalebox{0.6}{±0.62}}$} &  63.85$^{\scalebox{0.6}{±0.62}}$ &           63.78$^{\scalebox{0.6}{±0.64}}$ &  63.96$^{\scalebox{0.6}{±0.60}}$ \\
& UNI$\rightarrow$UNI & \cmark &  63.96$^{\scalebox{0.6}{±0.59}}$ &           63.68$^{\scalebox{0.6}{±0.67}}$ &  \textbf{63.69$^{\scalebox{0.6}{±0.68}}$} &  63.69$^{\scalebox{0.6}{±0.68}}$ &           63.68$^{\scalebox{0.6}{±0.66}}$ &  63.98$^{\scalebox{0.6}{±0.60}}$ \\

\hline
\multirow{24}{*}{ \rotatebox[origin=c]{90}{Places365}} 
& D3$\rightarrow$D1 & \xmark &  60.04$^{\scalebox{0.6}{±0.20}}$ &  \textbf{66.01$^{\scalebox{0.6}{±0.16}}$} &           65.71$^{\scalebox{0.6}{±0.15}}$ &  64.64$^{\scalebox{0.6}{±0.88}}$ &           66.00$^{\scalebox{0.6}{±0.12}}$ &  66.75$^{\scalebox{0.6}{±0.14}}$ \\
& D3$\rightarrow$D1 & \cmark &  60.04$^{\scalebox{0.6}{±0.20}}$ &  \textbf{66.03$^{\scalebox{0.6}{±0.16}}$} &           65.74$^{\scalebox{0.6}{±0.15}}$ &  64.66$^{\scalebox{0.6}{±0.92}}$ &           65.96$^{\scalebox{0.6}{±0.16}}$ &  66.73$^{\scalebox{0.6}{±0.14}}$ \\
& D3$\rightarrow$D2 & \xmark &  54.11$^{\scalebox{0.6}{±0.08}}$ &  \textbf{54.80$^{\scalebox{0.6}{±0.07}}$} &           54.24$^{\scalebox{0.6}{±0.08}}$ &  53.26$^{\scalebox{0.6}{±0.78}}$ &           53.96$^{\scalebox{0.6}{±0.08}}$ &  55.16$^{\scalebox{0.6}{±0.05}}$ \\
& D3$\rightarrow$D2 & \cmark &  54.11$^{\scalebox{0.6}{±0.08}}$ &  \textbf{54.79$^{\scalebox{0.6}{±0.06}}$} &           54.23$^{\scalebox{0.6}{±0.07}}$ &  53.22$^{\scalebox{0.6}{±0.81}}$ &           53.87$^{\scalebox{0.6}{±0.06}}$ &  55.17$^{\scalebox{0.6}{±0.05}}$ \\
& D3$\rightarrow$D3 & \xmark &  50.84$^{\scalebox{0.6}{±0.06}}$ &  \textbf{50.87$^{\scalebox{0.6}{±0.08}}$} &           50.11$^{\scalebox{0.6}{±0.12}}$ &  48.67$^{\scalebox{0.6}{±1.01}}$ &           49.30$^{\scalebox{0.6}{±0.12}}$ &  50.99$^{\scalebox{0.6}{±0.05}}$ \\
& D3$\rightarrow$D3 & \cmark &  50.84$^{\scalebox{0.6}{±0.06}}$ &  \textbf{50.76$^{\scalebox{0.6}{±0.09}}$} &           50.08$^{\scalebox{0.6}{±0.12}}$ &  48.60$^{\scalebox{0.6}{±1.04}}$ &           49.22$^{\scalebox{0.6}{±0.10}}$ &  50.99$^{\scalebox{0.6}{±0.05}}$ \\
& D3$\rightarrow$D4 & \xmark &  49.66$^{\scalebox{0.6}{±0.05}}$ &  \textbf{49.76$^{\scalebox{0.6}{±0.05}}$} &           49.13$^{\scalebox{0.6}{±0.13}}$ &  48.27$^{\scalebox{0.6}{±0.35}}$ &           48.37$^{\scalebox{0.6}{±0.08}}$ &  49.97$^{\scalebox{0.6}{±0.08}}$ \\
& D3$\rightarrow$D4 & \cmark &  49.66$^{\scalebox{0.6}{±0.05}}$ &  \textbf{49.77$^{\scalebox{0.6}{±0.05}}$} &           49.10$^{\scalebox{0.6}{±0.13}}$ &  48.21$^{\scalebox{0.6}{±0.37}}$ &           48.31$^{\scalebox{0.6}{±0.08}}$ &  49.98$^{\scalebox{0.6}{±0.07}}$ \\
& D3$\rightarrow$D5 & \xmark &  45.20$^{\scalebox{0.6}{±0.08}}$ &  \textbf{49.68$^{\scalebox{0.6}{±0.14}}$} &           48.90$^{\scalebox{0.6}{±0.20}}$ &  47.32$^{\scalebox{0.6}{±0.93}}$ &           48.09$^{\scalebox{0.6}{±0.02}}$ &  50.28$^{\scalebox{0.6}{±0.14}}$ \\
& D3$\rightarrow$D5 & \cmark &  45.19$^{\scalebox{0.6}{±0.07}}$ &  \textbf{49.68$^{\scalebox{0.6}{±0.15}}$} &           48.87$^{\scalebox{0.6}{±0.22}}$ &  47.24$^{\scalebox{0.6}{±0.99}}$ &           48.11$^{\scalebox{0.6}{±0.03}}$ &  50.31$^{\scalebox{0.6}{±0.14}}$ \\
& D3$\rightarrow$UNI & \xmark &  47.10$^{\scalebox{0.6}{±0.03}}$ &  \textbf{47.95$^{\scalebox{0.6}{±0.06}}$} &           47.64$^{\scalebox{0.6}{±0.13}}$ &  46.45$^{\scalebox{0.6}{±0.95}}$ &           46.87$^{\scalebox{0.6}{±0.04}}$ &  48.64$^{\scalebox{0.6}{±0.03}}$ \\
& D3$\rightarrow$UNI & \cmark &  47.10$^{\scalebox{0.6}{±0.03}}$ &  \textbf{47.94$^{\scalebox{0.6}{±0.06}}$} &           47.62$^{\scalebox{0.6}{±0.13}}$ &  46.38$^{\scalebox{0.6}{±0.98}}$ &           46.83$^{\scalebox{0.6}{±0.04}}$ &  48.65$^{\scalebox{0.6}{±0.03}}$ \\
& UNI$\rightarrow$D1 & \xmark &  53.12$^{\scalebox{0.6}{±0.09}}$ &  \textbf{67.29$^{\scalebox{0.6}{±0.14}}$} &           66.97$^{\scalebox{0.6}{±0.18}}$ &  66.71$^{\scalebox{0.6}{±0.24}}$ &           67.21$^{\scalebox{0.6}{±0.16}}$ &  68.10$^{\scalebox{0.6}{±0.11}}$ \\
& UNI$\rightarrow$D1 & \cmark &  53.12$^{\scalebox{0.6}{±0.09}}$ &  \textbf{67.29$^{\scalebox{0.6}{±0.14}}$} &           66.97$^{\scalebox{0.6}{±0.18}}$ &  66.72$^{\scalebox{0.6}{±0.24}}$ &           67.21$^{\scalebox{0.6}{±0.16}}$ &  68.10$^{\scalebox{0.6}{±0.11}}$ \\
& UNI$\rightarrow$D2 & \xmark &  52.29$^{\scalebox{0.6}{±0.06}}$ &  \textbf{56.95$^{\scalebox{0.6}{±0.08}}$} &           56.56$^{\scalebox{0.6}{±0.09}}$ &  56.52$^{\scalebox{0.6}{±0.11}}$ &           56.88$^{\scalebox{0.6}{±0.09}}$ &  57.38$^{\scalebox{0.6}{±0.10}}$ \\
& UNI$\rightarrow$D2 & \cmark &  52.29$^{\scalebox{0.6}{±0.06}}$ &  \textbf{56.95$^{\scalebox{0.6}{±0.07}}$} &           56.56$^{\scalebox{0.6}{±0.09}}$ &  56.51$^{\scalebox{0.6}{±0.11}}$ &           56.89$^{\scalebox{0.6}{±0.09}}$ &  57.37$^{\scalebox{0.6}{±0.11}}$ \\
& UNI$\rightarrow$D3 & \xmark &  52.10$^{\scalebox{0.6}{±0.07}}$ &  \textbf{53.50$^{\scalebox{0.6}{±0.07}}$} &           53.06$^{\scalebox{0.6}{±0.08}}$ &  53.01$^{\scalebox{0.6}{±0.07}}$ &           53.37$^{\scalebox{0.6}{±0.10}}$ &  53.72$^{\scalebox{0.6}{±0.08}}$ \\
& UNI$\rightarrow$D3 & \cmark &  52.10$^{\scalebox{0.6}{±0.07}}$ &  \textbf{53.50$^{\scalebox{0.6}{±0.07}}$} &           53.06$^{\scalebox{0.6}{±0.09}}$ &  53.01$^{\scalebox{0.6}{±0.07}}$ &           53.37$^{\scalebox{0.6}{±0.10}}$ &  53.72$^{\scalebox{0.6}{±0.08}}$ \\
& UNI$\rightarrow$D4 & \xmark &  51.68$^{\scalebox{0.6}{±0.10}}$ &  \textbf{52.61$^{\scalebox{0.6}{±0.07}}$} &           52.30$^{\scalebox{0.6}{±0.13}}$ &  52.28$^{\scalebox{0.6}{±0.13}}$ &           52.51$^{\scalebox{0.6}{±0.08}}$ &  52.81$^{\scalebox{0.6}{±0.06}}$ \\
& UNI$\rightarrow$D4 & \cmark &  51.68$^{\scalebox{0.6}{±0.10}}$ &  \textbf{52.61$^{\scalebox{0.6}{±0.08}}$} &           52.29$^{\scalebox{0.6}{±0.13}}$ &  52.28$^{\scalebox{0.6}{±0.13}}$ &           52.51$^{\scalebox{0.6}{±0.08}}$ &  52.81$^{\scalebox{0.6}{±0.05}}$ \\
& UNI$\rightarrow$D5 & \xmark &  52.49$^{\scalebox{0.6}{±0.15}}$ &  \textbf{53.78$^{\scalebox{0.6}{±0.06}}$} &           53.31$^{\scalebox{0.6}{±0.15}}$ &  53.23$^{\scalebox{0.6}{±0.15}}$ &           53.59$^{\scalebox{0.6}{±0.10}}$ &  54.11$^{\scalebox{0.6}{±0.10}}$ \\
& UNI$\rightarrow$D5 & \cmark &  52.49$^{\scalebox{0.6}{±0.14}}$ &  \textbf{53.78$^{\scalebox{0.6}{±0.06}}$} &           53.31$^{\scalebox{0.6}{±0.14}}$ &  53.23$^{\scalebox{0.6}{±0.15}}$ &           53.59$^{\scalebox{0.6}{±0.09}}$ &  54.11$^{\scalebox{0.6}{±0.10}}$ \\
& UNI$\rightarrow$UNI & \xmark &  51.77$^{\scalebox{0.6}{±0.09}}$ &  \textbf{51.89$^{\scalebox{0.6}{±0.07}}$} &           51.58$^{\scalebox{0.6}{±0.10}}$ &  51.56$^{\scalebox{0.6}{±0.07}}$ &           51.75$^{\scalebox{0.6}{±0.07}}$ &  52.01$^{\scalebox{0.6}{±0.09}}$ \\
& UNI$\rightarrow$UNI & \cmark &  51.77$^{\scalebox{0.6}{±0.09}}$ &  \textbf{51.89$^{\scalebox{0.6}{±0.08}}$} &           51.58$^{\scalebox{0.6}{±0.10}}$ &  51.56$^{\scalebox{0.6}{±0.07}}$ &           51.75$^{\scalebox{0.6}{±0.06}}$ &  52.01$^{\scalebox{0.6}{±0.08}}$ \\

\hline
\end{tabular}
\vspace*{.5mm}
\caption{\text{``Estimate test priors or directly the prior ratio?''} Accuracy (± std. dev.) after adaptation with the priors estimated by SCM$^\text{L}$ or with the prior ratio estimated by BBSE\cite{lipton2018detecting}. Results on CIFAR are averaged from 10 experiments, results on Places and ImageNet are averaged from 5 experiments. Best results are displayed in bold.}
\label{tab:prior_ratio_noretrain_sup} 
\end{table*}

\end{document}